\documentclass[runningheads]{llncs}
\usepackage{amsmath,amsfonts,bm}

\def\eqref#1{equation~\ref{#1}}

\def\1{\bm{1}}

\DeclareMathAlphabet{\mathsfit}{\encodingdefault}{\sfdefault}{m}{sl}
\SetMathAlphabet{\mathsfit}{bold}{\encodingdefault}{\sfdefault}{bx}{n}

\newcommand{\R}{\mathbb{R}}

\usepackage{graphicx}
\def\bs{\bm}

\usepackage{epsfig}
\usepackage{amssymb}
\usepackage{amsmath,amsfonts,bm}
\usepackage{multirow}
\usepackage{subfigure}
\usepackage{booktabs}
\usepackage{array}
\usepackage{wrapfig}
\usepackage{algorithm, algorithmic}
\usepackage[flushleft]{threeparttable}
\usepackage[misc]{ifsym}
\begin{document}
\title{Generative Max-Mahalanobis Classifiers for Image Classification, Generation and More}

\titlerunning{GMMC for Image Classification, Generation and More}
%
\author{Xiulong Yang, Hui Ye, Yang Ye, Xiang Li \& Shihao Ji{\Letter}  \\
}

%
\authorrunning{Xiulong Yang, Hui Ye, Yang Ye, Xiang Li \& Shihao Ji}
%
\institute{Department of Computer Science\\
Georgia State University \\
\email{\{xyang22,hye2,yye10,xli62,sji\}@gsu.edu}}

\tocauthor{Xiulong Yang, Hui Ye, Yang Ye, Xiang Li \& Shihao Ji}
\toctitle{Generative Max-Mahalanobis Classifiers for Image Classification, Generation and More}

\maketitle              

\begin{abstract}
Joint Energy-based Model (JEM) of ~\cite{jem} shows that a standard softmax classifier can be reinterpreted as an energy-based model (EBM) for the joint distribution $p(\bm{x}, y)$; the resulting model can be optimized  to improve calibration, robustness and out-of-distribution detection, while generating samples rivaling the quality of recent GAN-based approaches. However, the softmax classifier that JEM exploits is inherently discriminative and its latent feature space is not well formulated as probabilistic distributions, which may hinder its potential for image generation and incur training instability.  We hypothesize that generative classifiers, such as Linear Discriminant Analysis (LDA), might be more suitable for image generation since generative classifiers model the data generation process explicitly. This paper therefore investigates an LDA classifier for image classification and generation. In particular, the Max-Mahalanobis Classifier (MMC)~\cite{Pang2020Rethinking}, a special case of LDA, fits our goal very well.  We show that our Generative MMC (GMMC) can be trained discriminatively, generatively or jointly for image classification and generation. Extensive experiments on multiple datasets show that GMMC achieves state-of-the-art discriminative and generative performances, while outperforming JEM in calibration, adversarial robustness and out-of-distribution detection by a significant margin.
Our source code is available at \url{https://github.com/sndnyang/GMMC}.


\end{abstract}

\section{Introduction}

Over the past few years, deep neural networks (DNNs) have achieved state-of-the-art performance on a wide range of learning tasks, such as image classification, object detection, segmentation and image captioning~\cite{Krizhevsky2012,resnet16}. All of these breakthroughs, however, are achieved in the framework of discriminative models, which are known to be exposed to several critical issues, such as adversarial examples~\cite{advexample15}, calibration of uncertainty~\cite{guo2017calibration} and out-of-distribution detection~\cite{HenGim16}. Prior works have shown that generative training is beneficial to these models and can alleviate some of these issues at certain levels~\cite{ssl09,dempster1977maximum}. Yet, most recent research on generative models focus primarily on qualitative sample quality~\cite{biggan2018largeGAN,santurkar2019image,vahdat2020NVAE}, and the discriminative performances of state-of-the-art generative models are still far behind discriminative ones~\cite{behrmann2018invertible,chen2019residual,du2019implicit}.

\begin{figure}[t]

    \centering
    \subfigure[Softmax Classifier]{
        \includegraphics[width=0.38\columnwidth]{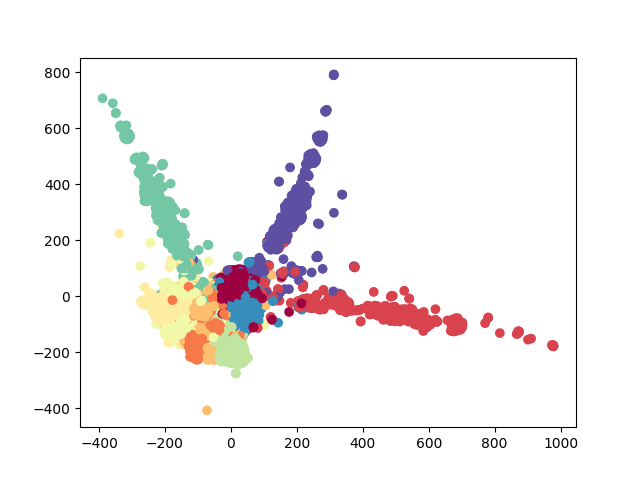}
        \label{figure:base_tsne}
    }
    \subfigure[JEM]{
        \includegraphics[width=0.38\columnwidth]{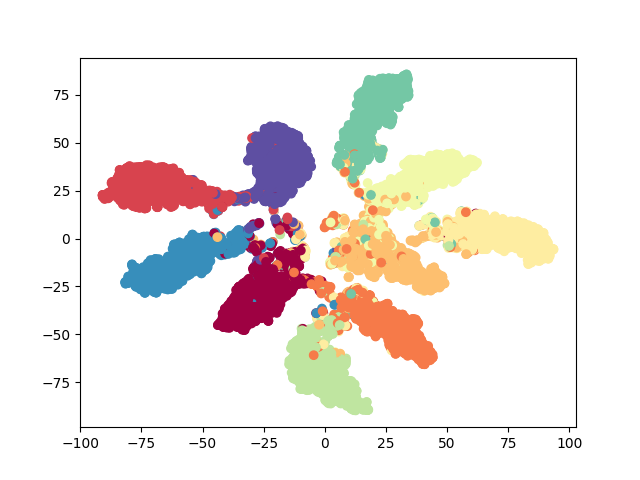}
        \label{figure:jem_tsne}
    }
    \subfigure[GMMC (Dis)]{
        \includegraphics[width=0.38\columnwidth]{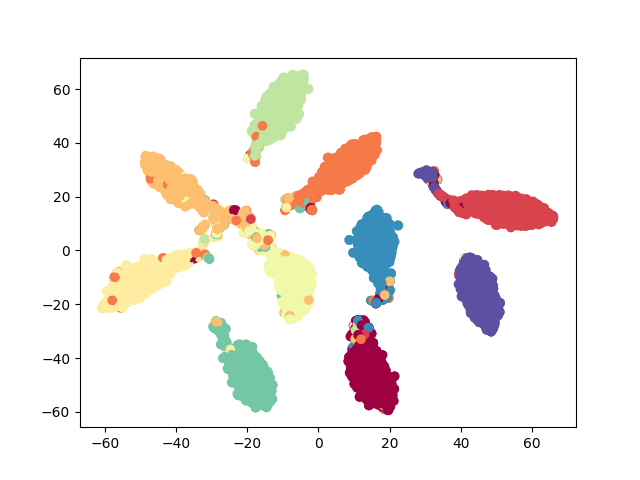}
        \label{figure:gmmc_dis_tsne}
    }
    \subfigure[GMMC (Gen)]{
        \includegraphics[width=0.38\columnwidth]{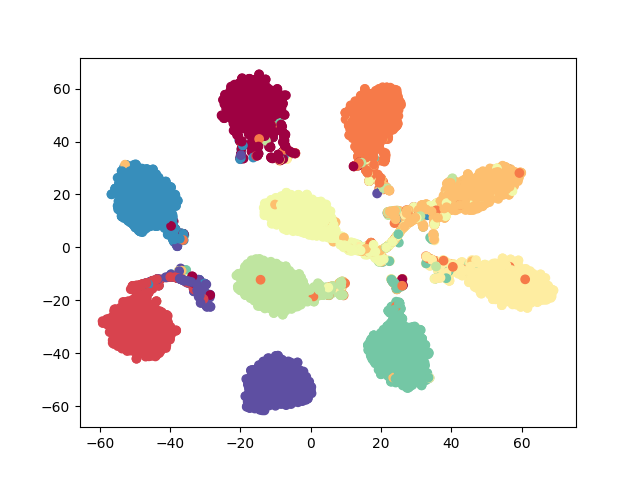}
        \label{figure:gmmc_gen_tsne}
    }
    \caption{t-SNE visualization of the latent feature spaces learned by different models trained on CIFAR10.}
    \label{figure:CIFAR10_tsne}

\end{figure}

Recently, there is a flurry of interest in closing the performance gap between generative models and discriminative models~\cite{du2019implicit,jem,Gao2020FlowEBM,IBINN20}. Among them, IGEBM~\cite{du2019implicit} and JEM~\cite{jem} are the two most representative ones, which reinterpret CNN classifiers as the energy-based models (EBMs) for image generation. Since the CNN classifier is the only trained model, which has a high compositionality, it is possible that a single trained CNN model may encompass the generative capabilities into the discriminative model without sacrificing its discriminative power. Their works realize the potential of EBMs in hybrid modeling and achieve improved performances on discriminative and generative tasks. Specifically, JEM~\cite{jem} reinterprets the standard softmax classifier as an EBM and achieves impressive performances in image classification and generation simultaneously, and ignites a series of follow-up works \cite{ZhaJacGra20,grathwohl2020SteinEBM,Gao2020FlowEBM}.

However, the softmax classifier that JEM exploits is inherently discriminative, which may hinder its potential in image generation. To investigate this, we visualize the latent feature spaces learnt by a standard softmax classifier and by JEM through t-SNE~\cite{tsne2008} in Figs.~\ref{figure:base_tsne} and~\ref{figure:jem_tsne}, respectively. Apparently, the feature space of the softmax classifier has been improved significantly by JEM as manifested by higher inter-class separability and intra-class compactness. However, JEM's latent space is not well formulated as probabilistic distributions, which may limit its generative performance and incur training instability as observed in~\cite{jem}. We hypothesize that generative classifiers (e.g., LDA) might be more suitable for image classification and generation. This is because generative classifiers model the data generation process explicitly with probabilistic distributions, such as mixture of Gaussians, which aligns well with the generative process of image synthesis. Therefore, in this paper we investigate an LDA classifier for image classification and generation. In particular, the Max-Mahalanobis Classifier (MMC)~\cite{Pang2020Rethinking}, a special case of LDA, fits our goal very well since MMC formulates the latent feature space explicitly as the Max-Mahalanobis distribution~\cite{pang2018max}. Distinct to~\cite{Pang2020Rethinking}, we show that MMC can be trained discriminatively, generatively or jointly as an EBM. We term our algorithm Generative MMC (GMMC) given that it is a hybrid model for image classification and generation, while the original MMC~\cite{Pang2020Rethinking} is only for classification. 

As a comparison, Figs.~\ref{figure:gmmc_dis_tsne} and~\ref{figure:gmmc_gen_tsne} illustrate the latent feature spaces of GMMC optimized with discriminative training and generative training (to be discussed in Sec.~\ref{sec:method}), respectively. It can be observed that the latent feature spaces of GMMC are improved even further over that of JEM's with higher inter-class separability and intra-class compactness. Furthermore, the explicit generative modeling of GMMC leads to many auxiliary benefits, such as adversarial robustness, calibration of uncertainty and out-of-distribution detection, which will be demonstrated in our experiments. Our main contributions can be summarized as follows:

\begin{enumerate}
\item We introduce GMMC, a hybrid model for image classification and generation. As an alternative to the softmax classifier utilized in JEM, GMMC has a well-formulated latent feature distribution, which fits well with the generative process of image synthesis.
\item We show that GMMC can be trained discriminatively,  generatively or jointly with reduced complexity and improved stability as compared to JEM.
\item Our model matches or outperforms prior state-of-the-art hybrid models on multiple discriminative and generative tasks, including image classification, image synthesis, calibration of uncertainty, out-of-distribution detection and adversarial robustness.
\end{enumerate}

\section{Background and Related Work}

\subsection{Energy-based Models}

Energy-based models (EBMs)~\cite{lecun2006tutorial} define an energy function that assigns low energy values to samples drawn from data distribution and high values otherwise, such that any probability density $p_{\bs{\theta}}(\bs{x})$ can be expressed via a Boltzmann distribution as
\begin{equation}\label{eq:ebm_define}
  p_{\bs{\theta}}(\bs{x})=\exp \left(-E_{\bs{\theta}}(\bs{x})\right)/Z(\bs{\theta}),
\end{equation}
where $E_{\bs{\theta}}(\bs{x})$ is an energy function that maps each input $\bs{x}\in\mathbb{R}^D$ to a scalar, and $Z(\bs{\theta})$ is the normalizing constant (also known as the partition function) such that $p_{\bs{\theta}}(\bs{x})$ is a valid density function. 

The key challenge of training EBMs lies in estimating the partition function $Z(\bs{\theta})$, which is notoriously  intractable. 
The standard maximum likelihood estimation of parameters $\bs{\theta}$ is not straightforward, and a number of sampling-based approaches have been proposed to approximate it effectively. Specifically, the gradient of the log-likelihood of a single sample $\bs{x}$ w.r.t. $\bs{\theta}$ can be expressed as
\begin{align}\label{eq:log}
  \frac{\partial\log p_{\bs{\theta}}(\bs{x})}{\partial\bs{\theta}}=\mathbb{E}_{p_{\bs{\theta}}(\bs{x}')}\frac{\partial E_{\bs{\theta}}(\bs{x}')}{\partial\bs{\theta}}-\frac{\partial E_{\bs{\theta}}(\bs{x})}{\partial\bs{\theta}},
\end{align}
where the expectation is over model distribution $p_{\bs{\theta}}(\bs{x}')$, sampling from which is challenging due to the intractable $Z(\bs{\theta})$. Therefore, MCMC and Gibbs sampling~\cite{hinton2002training} have been proposed previously to estimate the expectation efficiently. To speed up the mixing for effective sampling, recenlty Stochastic Gradient Langevin Dynamics (SGLD)~\cite{welling2011bayesian} has been used to train EBMs by exploiting the gradient information~\cite{nijkamp2019learning,du2019implicit,jem}. Specifically, to sample from $p_{\bs{\theta}}(\bs{x})$, SGLD follows
\begin{align}\label{eq:sgld}
&x_0\sim p_0(\bs{x}), \;
&x_{i+1}=x_i-\frac{\alpha}{2} \frac{\partial E_{\bs{\theta}}(\bs{x}_i)}{\partial \bs{x}_i} + \alpha\epsilon, \;\; \epsilon \sim \mathcal{N} (0,1),
\end{align}
where $p_0(\bs{x})$ is typically a uniform distribution over $[-1,1]$, whose samples are refined via noisy gradient decent with step-size $\alpha$, which should be
decayed following a polynomial schedule.

Besides JEM~\cite{jem} that we discussed in the introduction,~\cite{xie2016theory} is an earlier work that derives a generative CNN model from the commonly used discriminative CNN by treating it as an EBM, where
the authors factorize the loss function $\log p(\bs{x}|y)$ as an EBM. Following-up works, such as \cite{nijkamp2019anatomy,du2019implicit}, scale the training of EBMs to high-dimensional data using SGLD. However, all of these previous methods define $p(\bs{x}|y)$ or $p(\bs{x})$ as an EBM, while our GMMC defines an EBM on $p(\bs{x},y)$ by following a mixture of Gaussian distribution, which simplifies the maximum likelihood estimation and achieves improved performances in many discriminative and generative tasks.

\subsection{Alternatives to the Softmax Classifier}

Softmax classifier has been widely used in state-of-the-art models for discriminative tasks due to its simplicity and efficiency. However, softmax classifier is known particularly vulnerable to adversarial attacks because the latent feature space induced by softmax classifier is typically not well separated (as shown in Fig.~\ref{figure:CIFAR10_tsne}(a)). Some recent works propose to use generative classifiers to better formulate the latent space distributions in order to improve its robustness to adversarial examples. For example, Wan et al.~\cite{wan2018rethinking} propose to model the latent feature space as mixture of Gaussians and encourages stronger intra-class compactness and larger inter-class separability by introducing large margins between classes. Different from~\cite{wan2018rethinking}, Pang et al.~\cite{pang2018max} pre-design the centroids based on the Max-Mahalanobis distribution (MMD), other than learning them from data. The authors prove that if the latent feature space distributes as an MMD, the LDA classifier will have the best robustness to adversarial examples. Taking advantage of the benefits of MMD, Pang et al.~\cite{Pang2020Rethinking} further propose a max-Mahalanobis center regression loss, which induces much denser feature regions and improves the robustness of trained models. Compared with softmax classifier, all these works can generate better latent feature spaces to improve the robustness of models for the task of classification. Our GMMC is built on the basic framework of MMC, but we reinterpret MMC as an EBM for image classification and generation. Moreover, we show that the generative training of MMC can further improve calibration, adversarial robustness and out-of-distribution detection.

\section{Methodology}\label{sec:method}

We assume a Linear Discriminant Analysis (LDA) classifier is defined as: $\bs{\phi}(\bs{x})$, $\bs{\mu}=\{\bs{\mu}_y, y=1,2,\cdots,C\}$ and $\bs{\pi}=\{\pi_y=\frac{1}{C}, y=1,2,\cdots,C\}$ for $C$-class classification, where $\bs{\phi}(\bs{x})\in\mathbb{R}^d$ is the feature representation of $\bs{x}$ extracted by a CNN, parameterized by $\bs{\phi}$\footnote{To avoid notational clutter in later derivations, we use $\bs{\phi}$ to denote a CNN feature extractor and its parameter. But the meaning of $\bs{\phi}$ is clear given the context.}, and $\bs{\mu}_y\in\mathbb{R}^d$ is the mean of a Gaussians distribution with a diagonal covariance matrix $\gamma^2\bs{I}$, i.e., $p_{\bs{\theta}}(\bs{\phi}(\bs{x})|y)=\mathcal{N}(\bs{\mu}_y, \gamma^2\bs{I})$. Therefore, we can parameterize LDA by $\bs{\theta}=\{\bs{\phi}, \bs{\mu}\}$\footnote{We can treat $\gamma$ as a tunable hyperparameter or we can estimate it by post-processing. In this work, we take the latter approach as discussed in Sec.~\ref{sec:disc}.}. Instead of using this regular LDA classifier, in this paper the max-Mahalanobis classifier (MMC)~\cite{Pang2020Rethinking}, a special case of LDA, is considered. Different from the LDA modeling above, in MMC $\bs{\mu}=\{\bs{\mu}_y, y=1,2,\cdots,C\}$ is pre-designed to induce compact feature representations for model robustness. We found that the MMC modeling fits our goal better than the regular LDA classifier due to its improved training stability and boosted adversarial robustness. Therefore, in the following we focus on the MMC modeling for image classification and generation. As such, the learnable parameters of MMC reduce to $\bs{\theta}=\{\bs{\phi}\}$, and the pseudo-code of calculating pre-designed $\bs{\mu}$ can be found in Algorithm 2 of the  appendix~\ref{app:mu}. 
Fig.~\ref{figure:lda_diagram} provides an overview of the training and test of our GMMC algorithm, with the details discussed below.

Instead of maximizing $p_{\bs{\theta}}(y|\bs{x})$ as in standard softmax classifier, following JEM~\cite{jem} we maximize the joint distribution $p_{\bs{\theta}}(\bs{x},y)$, which follows a mixture of Gaussians distribution in GMMC. To optimize $\log p_{\bs{\theta}}(\bs{x},y)$, we can consider three different approaches.

\begin{figure*}[ht]

    \centering
    \includegraphics[width=0.95\textwidth]{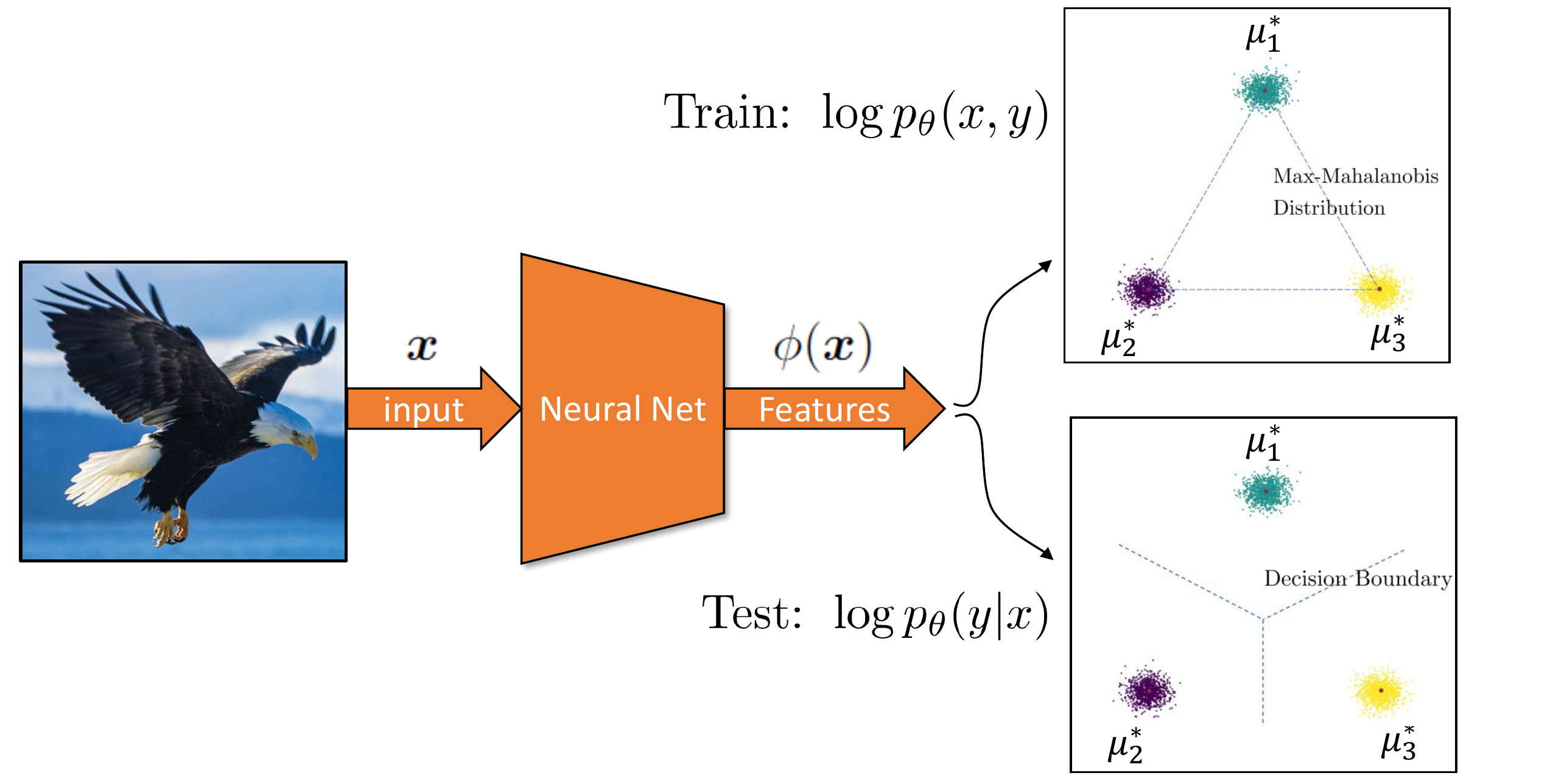}
    \caption{Overview of GMMC for training and test, where the model can be trained discriminatively, generatively or jointly. $\{\bs{\mu}_1^*,\bs{\mu}_2^*,\cdots,\bs{\mu}_C^*\}$ are pre-designed according to MMD~\cite{pang2018max}. Only $\bs{\theta}=\{\bs{\phi}\}$ is learned from data.}
    \label{figure:lda_diagram}

\end{figure*}

\subsection{Approach 1: Discriminative Training}\label{sec:disc}
\vspace{-5pt}
According to the MMC modeling above, the joint distribution $p_{\bs{\theta}}(\bs{x},y)$ can be expressed as
\begin{align}\label{eq:disc0}
  p_{\bs{\theta}}(\bs{x},y)&=p(y) p_{\bs{\theta}}(\bs{x}|y)\propto\frac{1}{C}(2\pi\gamma^2)^{-d/2}\exp(-\frac{1}{2\gamma^2}||\bs{\phi}(\bs{x})-\bs{\mu}_y||_2^2)\nonumber\\
  &=\frac{\exp(-\frac{1}{2\gamma^2}||\bs{\phi}(\bs{x})-\bs{\mu}_y||_2^2)}{Z(\bs{\theta})}=\frac{\exp(-E_{\bs{\theta}}(\bs{x},y))}{Z(\bs{\theta})}
\end{align}
where we define $E_{\bs{\theta}}(\bs{x},y)=\frac{1}{2\gamma^2}||\bs{\phi}(\bs{x})-\bs{\mu}_y||_2^2$, and $Z(\bs{\theta})=\int\exp(-E_{\bs{\theta}}(\bs{x},y))d\bs{x}dy$, which is an intractable partition function. To avoid the expense of evaluating the partition function, we follow Mnih and Teh~\cite{MniTeh12} and approximate $Z(\bs{\theta})$ as a constant (e.g., $Z(\bs{\theta})=1$). This turns out to be an effective approximation for neural networks with lots of parameters as it encourages the model to have ``self-normalized" outputs. With this approximation, the log of the joint distribution can be simplified as
\begin{align}\label{eq:disc}
  \log p_{\bs{\theta}}(\bs{x},y)&= -\frac{1}{2\gamma^2}||\bs{\phi}(\bs{x})-\bs{\mu}_y||_2^2 - \log Z(\bs{\bs{\theta}})\nonumber\\
  &\approx-E_{\bs{\theta}}(\bs{x},y) + \text{constant}.
\end{align}
To optimize the parameters $\bs{\theta}$, we can simply compute gradient of Eq.~\ref{eq:disc} w.r.t. $\bs{\theta}$, and update the parameters by stochastic gradient descent (SGD)~\cite{sgd}. Note that $\gamma$ is a constant in Eq.~\ref{eq:disc}, its effect can be absorbed into the learning rate when optimizing Eq.~\ref{eq:disc} via SGD. After convergence, we can estimate $\gamma^2=\frac{1}{d}\left(\frac{1}{N}\sum_{i=1}^N||\bs{\phi}(\bs{x}_i)-\bs{\mu}_i||_2^2\right)$ from training set by using optimized $\bs{\phi}$ and pre-designed $\bs{\mu}$.

Note that Eq.~\ref{eq:disc} boils down to the same objective that MMC~\cite{Pang2020Rethinking} proposes, i.e. the center regression loss. While MMC reaches to this objective from the perspective of inducing compact feature representations for model robustness, we arrive at this objective by simply following the principle of maximum likelihood estimation of model parameter $\bs{\theta}$ of joint density $p_{\bs{\theta}}(\bs{x},y)$ with the ``self-normalization" approximation~\cite{MniTeh12}.

\subsection{Approach 2: Generative Training}

Comparing Eq.~\ref{eq:disc0} with the definition of EBM (\ref{eq:ebm_define}), we can also treat the joint density
\begin{align}\label{eq:ebm}
  p_{\bs{\theta}}(\bs{x},y)=\frac{\exp(-\frac{1}{2\gamma^2}||\bs{\phi}(\bs{x})-\bs{\mu}_y||_2^2)}{Z(\bs{\theta})}
\end{align} 
as an EBM with $E_{\bs{\theta}}(\bs{x},y)=\frac{1}{2\gamma^2}||\bs{\phi}(\bs{x})-\bs{\mu}_y||_2^2$ defined as an energy function of $(\bs{x},y)$.

Following the maximum likelihood training of EBM (\ref{eq:log}), to optimize Eq.~\ref{eq:ebm}, we can compute its gradient w.r.t. $\bs{\theta}$ as
\begin{align}\label{eq:ebm_grad}
  \!\!\!\!\frac{\partial\log p_{\bs{\theta}}(\bs{x},y)}{\partial\bs{\theta}}\!=\!\beta\mathbb{E}_{p_{\bs{\theta}}(\bs{x}',y')}\!\frac{\partial E_{\bs{\theta}}(\bs{x}',y')}{\partial\bs{\theta}}\!-\!\frac{\partial E_{\bs{\theta}}(\bs{x},y)}{\partial\bs{\theta}},
\end{align}
where the expectation is over the joint density $p_{\bs{\theta}}(\bs{x}',y')$, sampling from which is challenging due to the intractable $Z(\bs{\theta})$, and $\beta$ is a hyperparameter that balances the contributions from the two terms. Different value of $\beta$ has a significant impact to the performance. From our experiments, we find that $\beta\!=\!0.5$ works very well in all our cases. Therefore, we set $\beta\!=\!0.5$ as the default value. Notably, the two terms of our GMMC (Eq.~\ref{eq:ebm_grad}) are both defined on the same energy function $E_{\bs{\theta}}(\bs{x},y)$, while the two terms of JEM~\cite{jem} are defined on $p(\bs{x})$ and $p(y|\bs{x})$, respectively, which are computationally more expensive and might incur training instability as we will discuss in Sec.~\ref{sec:joint}.



The tricky part is how to generate samples $(\bs{x}',y')\sim p_{\bs{\theta}}(\bs{x}',y')$ to estimate the first term of Eq.~\ref{eq:ebm_grad}. We can follow the mixture of Gaussians assumption of MMC. That is, $p_{\bs{\theta}}(\bs{x}',y')=p(y')p_{\bs{\theta}}(\bs{x}'|y')$: (1) sample $y'\sim p(y')=\frac{1}{C}$, and then (2) sample $\bs{x}'\sim p_{\bs{\theta}}(\bs{x}'|y')\propto\mathcal{N}(\bs{\mu}_{y'}, \gamma^2\bs{I})$. To sample $\bs{x}'$, again we can consider two choices.

\paragraph{\textbf{(1) Staged Sampling}}
We can first sample $\bs{z}_{\bs{x}'}\sim \mathcal{N}(\bs{\mu}_{y'}, \gamma^2\bs{I})$, and then find an $\bs{x}'$ to minimize $E_{\bs{\theta}}(\bs{x}')=\frac{1}{2\gamma^2}||\bs{\phi}(\bs{x}')-\bs{z}_{\bs{x}'}||_2^2$. This can be achieved by
\begin{align}\label{eq:ebm_staged}
\bs{x}'_0\sim p_0(\bs{x}'),\;\;\;\;\;\;\;\bs{x}'_{t+1}=\bs{x}'_t-\alpha \frac{\partial E_{\bs{\theta}}(\bs{x}'_t)}{\partial \bs{x}'_t},
\end{align}
where $p_0(\bs{x})$ is typically a uniform distribution over $[-1,1]$. Note that this is similar to SGLD (see Eq.~\ref{eq:sgld}) but without a noisy term. Thus, the training could be more stable. In addition, the function $E_{\bs{\theta}}(\bs{x}')=\frac{1}{2\gamma^2}||\bs{\phi}(\bs{x}')-\bs{z}_{\bs{x}'}||_2^2$ is just an $L_2$ regression loss (not an LogSumExp function as used in JEM~\cite{jem}). This may lead to additional numerical stability.

\paragraph{\textbf{(2) Noise Injected Sampling}}
We can first sample $\bs{z}\sim \mathcal{N}(\bs{0}, \bs{I})$, then by the reparameterization trick we have $\bs{z}_{\bs{x}'}=\gamma \bs{z} + \bs{\mu}_{y'}$. Finally, we can find an $\bs{x}'$ to minimize 
\begin{align}\label{eq:ebm_noisy_obj}
&E_{\bs{\theta}}(\bs{x}')\!=\!\frac{1}{2\gamma^2}||\bs{\phi}(\bs{x}')-\bs{z}_{\bs{x}'}||_2^2\!=\!\frac{1}{2\gamma^2}||\bs{\phi}(\bs{x}')-\bs{\mu}_{y'}-\gamma \bs{z}||_2^2\nonumber\\
&=\!\frac{1}{2\gamma^2}||\bs{\phi}(\bs{x}')\!-\!\bs{\mu}_{y'}||_2^2\!+\!\frac{1}{2}||\bs{z}||_2^2-\frac{1}{\gamma^2}<\!\!\bs{\phi}(\bs{x}')\!-\!\bs{\mu}_{y'},\!\gamma\bs{z}\!\!>\nonumber\\
&=\!E_{\bs{\theta}}(\bs{x}',y')\!+\!\frac{1}{2}||\bs{z}||_2^2-\frac{1}{\gamma}<\!\!\bs{\phi}(\bs{x}')-\bs{\mu}_{y'},\bs{z}\!\!>.
\end{align}
This can be achieved by
\begin{align}\label{eq:ebm_noisy}
&\bs{x}'_0\sim p_0(\bs{x}'),\nonumber\\
&\bs{x}'_{t+1}=\bs{x}'_t-\alpha \frac{\partial E_{\bs{\theta}}(\bs{x}'_t)}{\partial \bs{x}'_t}
=\bs{x}'_t-\alpha\frac{\partial E_{\bs{\theta}}(\bs{x}'_t,y')}{\partial \bs{x}'_t}+\alpha\frac{1}{\gamma}<\frac{\partial\phi(\bs{x}'_t)}{\partial \bs{x}'_t},\bs{z}>,
\end{align}
where we sample a different $\bs{z}$ at each iteration. As a result, Eq.~\ref{eq:ebm_noisy} is an analogy of SGLD (see Eq.~\ref{eq:sgld}). The difference is instead of using an unit Gaussian noise $\epsilon\sim\mathcal{N}(0,1)$ as in SGLD, a gradient-modulated noise (the 3rd term) is applied.

\begin{algorithm}[t]
\caption{Generative training of GMMC: Given model parameter $\bs{\theta}=\{\bs{\phi}\}$, step-size $\alpha$, replay buffer $B$, number of steps $\tau$, reinitialization frequency $\rho$}
\label{algo:1}
\begin{algorithmic}[1]
\WHILE{not converged}
\STATE Sample $\bs{x}$ and $y$ from dataset $\mathcal{D}$
\STATE Sample $(\bs{x}'_0, y') \sim B$ with probability $1-\rho$,  else $\bs{x}'_0 \sim \mathcal{U}(-1, 1), y' \sim p(y') = \frac{1}{C}$

\STATE Sample $\bs{z}_{\bs{x}'} \sim \mathcal{N}(\bs{\mu}_{y'}, \gamma^2\bs{I})$ if staged sampling
\FOR{$t \in [1, 2, \cdots, \tau]$}
  \STATE Sample $\bs{z} \sim \mathcal{N}(\bs{0},\bs{I}), \bs{z}_{\bs{x}'} = \bs{\mu}_{y'} + \gamma \bs{z}$ if noise injected sampling
  \STATE $\bs{x}'_t = \bs{x}'_{t-1} - \alpha\frac{\partial E_{\bs{\theta}}(\bs{x}'_{t-1})}{\partial \bs{x}'_{t-1}}$ (Eq.~\ref{eq:ebm_staged})
\ENDFOR
\STATE Calculate gradient with Eq.~\ref{eq:ebm_grad} from $(\bs{x},y)$ and $(\bs{x}'_{\tau},y')$ for model update
\STATE Add / replace updated $(\bs{x}'_{\tau}, y')$ back to $B$
\ENDWHILE
\end{algorithmic}
\end{algorithm}

Algorithm~\ref{algo:1} provides the pseudo-code of the generative training of GMMC, which follows a similar design of IGEBM~\cite{du2019implicit} and JEM~\cite{jem} with a replay buffer $B$. For brevity, only one real sample $(\bs{x},y)\sim\mathcal{D}$ and one generated sample $(\bs{x}',y')\sim p_{\bs{\theta}}(\bs{x}',y')$ are used to optimize the parameter $\bs{\theta}$. It is straightforward to generalize the pseudo-code above to a mini-batch training, which is used in our experiments. Compared to JEM, GMMC needs no additional calculation of $p_{\bs{\theta}}(y|\bs{x})$ and thus has reduced computational complexity.

\vspace{-5pt}
\subsection{Approach 3: Joint Training}~\label{sec:joint}

Comparing Eq.~\ref{eq:disc} and Eq.~\ref{eq:ebm_grad}, we note that the gradient of Eq.~\ref{eq:disc} is just the second term of Eq.~\ref{eq:ebm_grad}. Hence, we can use (Approach 1) discriminative training to pretrain $\bs{\theta}$, and then finetune $\bs{\theta}$ by (Approach 2) generative training. The transition between the two can be achieved by scaling up $\beta$ from 0 to a predefined value (e.g., 0.5). Similar joint training strategy can be applied to train JEM as well. However, from our experiments we note that this joint training of JEM is extremely unstable. We hypothesize that this is likely because the two terms of JEM are defined on $p(\bs{x})$ and $p(y|\bs{x})$, respectively, while the two terms of our GMMC (Eq.~\ref{eq:ebm_grad}) are defined on the same energy function $E_{\bs{\theta}}(\bs{x},y)$. Hence, the learned model parameters from the two training stages are more compatible in GMMC than in JEM. We will demonstrate the training issues of JEM and GMMC when we present results.

\subsection{GMMC for Inference}
After training with one of the three approaches discussed above, we get the optimized GMMC parameters $\bs{\theta}=\{\bs{\phi}\}$, the pre-designed $\bs{\mu}$ and the estimated $\gamma^2=\frac{1}{d}\left(\frac{1}{N}\sum_{i=1}^N||\bs{\phi}(\bs{x}_i)-\bs{\mu}_i||_2^2\right)$ from training set. We can then calculate class probabilities for classification 
\begin{align}\label{eq:classification}
p_{\bs{\theta}}(y|\bs{x}) = \frac{\exp(-\frac{1}{2\gamma^2}||\bs{\phi}(\bs{x})-\bs{\mu}_y||_2^2)}{\sum_{y'}\exp(-\frac{1}{2\gamma^2}||\bs{\phi}(\bs{x})-\bs{\mu}_{y'}||_2^2)}.
\end{align}

\section{Experiments}

We evaluate the performance of GMMC on multiple discriminative and generative tasks, including image classification, image generation, calibration of uncertainty, out-of-distribution detection and adversarial robustness. Since GMMC is inspired largely by JEM~\cite{jem}, for a fair comparison, our experiments closely follow the settings provided in the source code of JEM\footnote{\url{https://github.com/wgrathwohl/JEM}}. All our experiments are performed with PyTorch on Nvidia RTX GPUs. Due to page limit, details of the experimental setup are relegated to Appendix~\ref{app:exp}.

\subsection{Hybrid Modeling}

We train GMMC on three benchmark datasets: CIFAR10, CIFAR100~\cite{Krizhevsky2012} and SVHN~\cite{svhn11}, and compare it to the state-of-the-art hybrid models, as well as standalone generative and discriminative models. Following the settings of JEM, we use the Wide-ResNet~\cite{wideresnet16} as the backbone CNN model for JEM and GMMC. To evaluate the quality of generated images, we adopt Inception Score (IS)~\cite{imprgan16} and Fr\'{e}chet Inception Distance (FID)~\cite{heusel2017gans} as the evaluation metrics.

\begin{table}[ht!]

\begin{minipage}{0.6\linewidth}

\caption{Hybrid Modeling Results on CIFAR10.}
\label{table:hybrid_results}
\begin{center}
\begin{threeparttable}
\begin{tabular}{c|c|ccc}
\toprule
Class  & Model  & Acc \% $\uparrow$ & IS $\uparrow$ & FID $\downarrow$ \\
\midrule
\multirow{5}{*}{Hybrid} & Residual Flow & 70.3 & 3.60  & 46.4 \\
                        & Glow          & 67.6 & 3.92 & 48.9 \\
                        & IGEBM         & 49.1 & 8.30  & 37.9 \\
                        & JEM           & 92.9 & \bf{8.76} & 38.4 \\
                        & GMMC (Ours) & \bf{94.08} & 7.24 & \bf{37.0} \\
\midrule
\multirow{2}{*}{Disc.} & WRN w/ BN  & 95.8 & N/A & N/A  \\
                        & WRN w/o BN & 93.6 & N/A & N/A  \\
                        & GMMC (Dis) & 94.3 & N/A & N/A  \\
\midrule
\multirow{3}{*}{Gen.}  & SNGAN  &  N/A & 8.59  & 25.5  \\
                        & NCSN   &  N/A & 8.91  & 25.3 \\
\bottomrule
\end{tabular}
\end{threeparttable}
\end{center}

\end{minipage}
\begin{minipage}[h]{0.4\linewidth}  

\begin{center}
\begin{threeparttable}
\caption{Test Accuracy (\%) on SVHN and CIFAR100.}\label{table:svhn_cifar100_acc}\vspace{-65pt}
\begin{tabular}{c|cc}
\toprule
Model  & SVHN & CIFAR100 \\
\midrule
Softmax       & 96.6 & 72.6 \\
JEM           & 96.7 & 72.2  \\
GMMC (Dis) & 97.1 & \bf{75.4}  \\
GMMC (Gen) & \bf{97.2} & 73.9  \\
\bottomrule
\end{tabular}
\end{threeparttable}
\end{center}

\end{minipage}

\end{table}

\vspace{-5pt}
The results on CIFAR10, CIFAR100 and SVHN are shown in Table~\ref{table:hybrid_results} and Table~\ref{table:svhn_cifar100_acc}, respectively. It can be observed that GMMC outperforms the state-of-the-art hybrid models in terms of accuracy (94.08\%) and FID score (37.0), while being slightly worse in IS score. Since no IS and FID scores are commonly reported on SVHN and CIFAR100, we present the classification accuracies and generated samples on these two benchmarks. Our GMMC models achieve 97.2\% and 73.9\% accuracy on SVHN and CIFAR100, respectively, outperforming the softmax classifier and JEM by notable margins. Example images generated by GMMC for CIFAR10 are shown in Fig.~\ref{figure:cifar10_samples}. Additional GMMC generated images for CIFAR100 and SVHN can be found in Appendix~\ref{app:samples}.

\begin{figure}[ht]
    \centering
    \subfigure[Unconditional Samples]{
        \includegraphics[width=0.47\columnwidth]{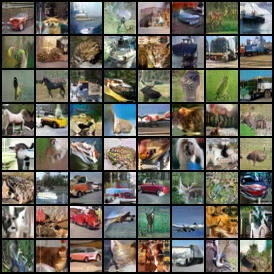}
        \label{figure:cifar10_un_sample}
    }
    \subfigure[Class-conditional Samples]{
        \includegraphics[width=0.47\columnwidth]{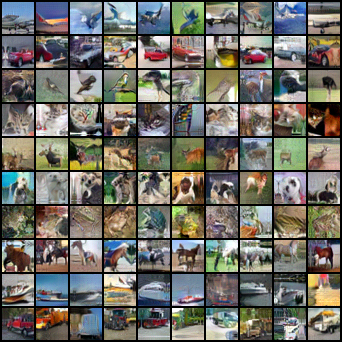}
        \label{figure:cifar10_con_sample}
    }

    \caption{Generated CIFAR10 Samples.}

    \label{figure:cifar10_samples}
\end{figure}

\subsection{Calibration}
\vspace{-5pt}
While modern deep models have grown more accurate in the past few years, recent researches have shown that their predictions could be over-confident~\cite{guo2017calibration}. Outputting an incorrect but confident decision can have catastrophic consequences. Hence, calibration of uncertainty for DNNs is a critical research topic. Here, the confidence is defined as $\max_y p(y|\bs{x})$ which is used to decide when to output a prediction. A well-calibrated, but less accurate model can be considerably more useful than a more accurate but less-calibrated model.

We train GMMC on the CIFAR10 dataset, and compare its Expected Calibration Error (ECE) score~\cite{guo2017calibration} to that of the standard softmax classifier and JEM. Results are shown in Fig.~\ref{figure:CIFAR10_cali} with additional results on SVHN and CIFAR100 provided in Appendix~\ref{app:calibration}. We find that the model trained by GMMC (Gen) achieves a much smaller ECE (1.33\% vs. 4.18\%), demonstrating GMMC's predictions are better calibrated than the competing methods.

\begin{figure}[ht!]

    \centering
    \subfigure[Softmax]{
        \includegraphics[width=0.225\columnwidth]{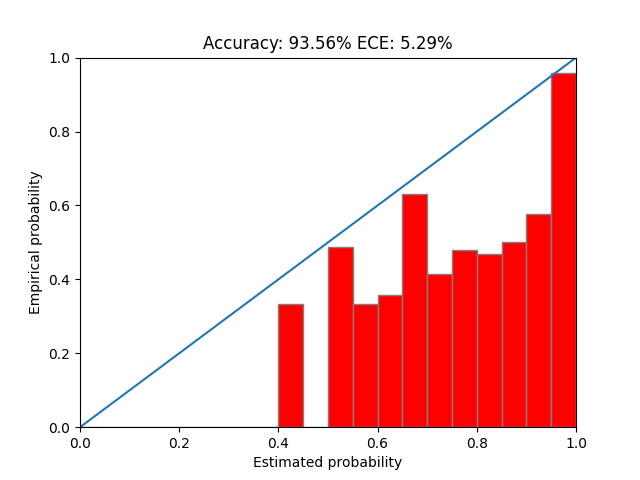}
        \label{figure:base_cali}
    }
    \subfigure[JEM]{
        \includegraphics[width=0.225\columnwidth]{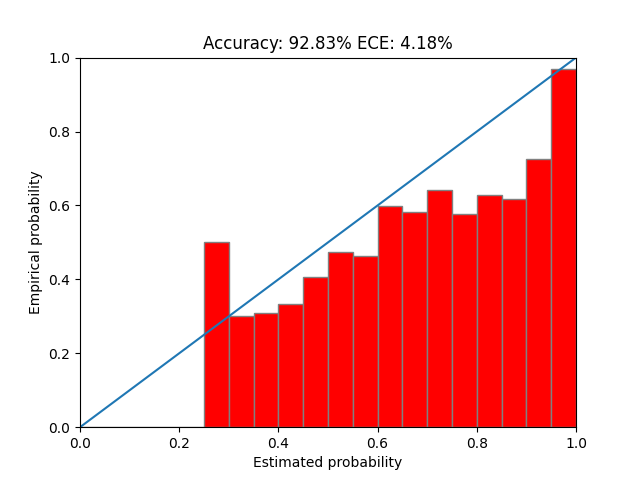}
        \label{figure:jem_cali}
    }
    \subfigure[GMMC (Dis)]{
        \includegraphics[width=0.225\columnwidth]{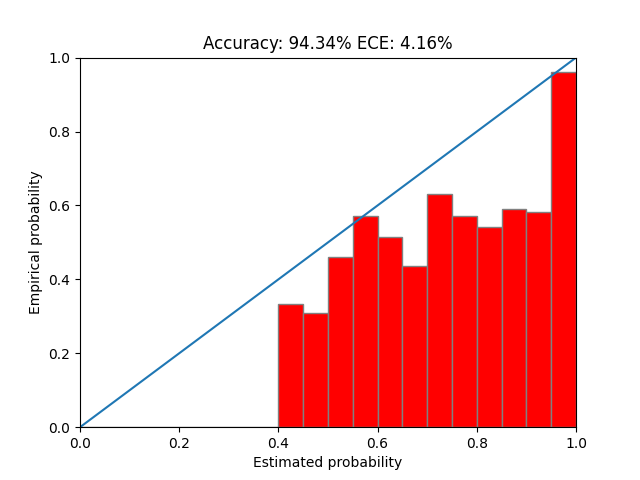}
        \label{figure:lda_dis_cali}
    }
    \subfigure[GMMC (Gen)]{
        \includegraphics[width=0.225\columnwidth]{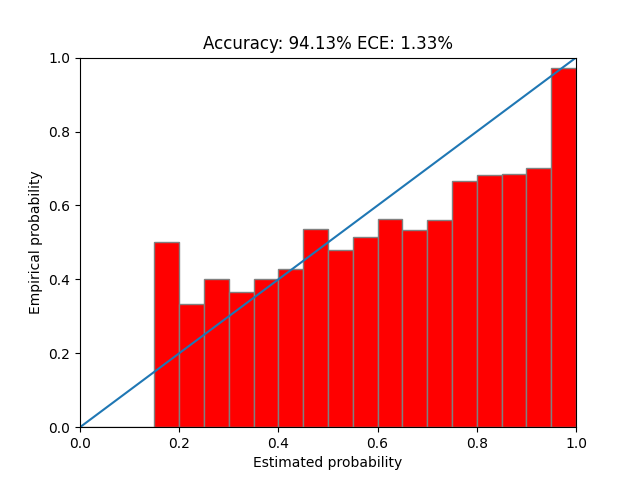}
        \label{figure:lda_gen_cali}
    }
    \caption{Calibration Results on CIFAR10. The smaller ECE is, the better.}
    \label{figure:CIFAR10_cali}

\end{figure}

\subsection{Out-Of-Distribution Detection}

The OOD detection is a binary classification problem, which outputs a score $s_{\bs{\theta}}(\bs{x}) \in \mathbb{R}$ for a given query $\bs{x}$. The model should be able to assign lower scores to OOD examples than to in-distribution examples, such that it can be used to distinguish two sets of examples. Following the settings of JEM~\cite{jem}, we use the Area Under the Receiver-Operating Curve (AUROC)~\cite{HenGim16} to evaluate the performance of OOD detection. In our experiments, three score functions are considered: the input density $p_{\bs{\theta}}(\bs{x})$~\cite{nalisnick2018deep}, the predictive distribution $p_{\bs{\theta}}(y|\bs{x})$~\cite{HenGim16}, and the approximate mass $\| \frac{\partial \log p_{\bs{\theta}}(\bs{x})}{\partial \bs{x}}\|$~\cite{jem}.

\paragraph{\textbf{(1) Input Density}}
A natural choice of $s_{\bs{\theta}}(\bs{x})$ is the input density $p_{\bs{\theta}}(\bs{x})$. For OOD detection, intuitively we consider examples with low $p(\bs{x})$ to be OOD. Quantitative results can be found in Table~\ref{cifar10_ood} (top). The corresponding distributions of scores are visualized in Table~\ref{table:logpx_hist}. The GMMC model assigns higher likelihoods to in-distribution data than to the OOD data, outperforming all the other models by a significant margin.

\paragraph{\textbf{(2) Approximate Mass}}
Recent work of~\cite{nalisnick2019detecting} has found that likelihood may not be enough for OOD detection in high-dimensional space. Real samples from a distribution form the area of high probability \textit{mass}. But a point may have a high density while the surrounding area has a very low density, which indicates the density can change rapidly around it and that point is likely not a sample from the real data distribution. Thus, the norm of gradient of the log-density will be large compared to examples in the area \textit{mass}. Based on this reasoning, Grathwohl et al. propose a new OOD score: $\bs{\theta}(\bs{x})=-\| \frac{\partial \log p_{\bs{\theta}}(\bs{x})}{\partial \bs{x}}\|_2$. Adopting this score function, we find that our model still outperforms the other competing methods (JEM and IGEBM), as shown in Table~\ref{cifar10_ood} (bottom).

\paragraph{\textbf{(3) Predictive Distribution}}
Another useful OOD score is the maximum probability from a classifier's predictive distribution: $s_{\bs{\theta}}(\bs{x}) = \max_y p_{\bs{\theta}}(y|\bs{x})$. Hence, OOD performance using this score is highly correlated with a model's classification accuracy. The results can be found in Table~\ref{cifar10_ood} (middle). Interestingly, with this score function, there is no clear winner over four different benchmarks consistently, while GMMC performs similarly to JEM in most of the cases.

\begin{table*}[ht]
\caption{OOD Detection Results. Models are trained on CIFAR10. Values are AUROC.}
\label{cifar10_ood}

\begin{center}
\begin{threeparttable}
\begin{tabular}{c|c|cccc}
\toprule
$s_{\bs{\theta}}(\bs{x})$  & Model   & SVHN & CIFAR10 Interp & CIFAR100 & CelebA \\
\midrule
\multirow{5}{*}{$\log p_{\bs{\theta}}(\bs{x})$} & Unconditional Glow          & .05 & .51 & .55 & .57 \\
                                  & Class-Conditional Glow      & .07 & .45 & .51 & .53 \\
                                  & IGEBM                       & .63 & .70 & .50 & .70 \\
                                  & JEM                         & .67 & .65 & .67 & .75 \\
                                  & GMMC (Gen)                  & \bf{.84} & \bf{.75} & \bf{.84} & \bf{.86} \\
\midrule
\multirow{5}{*}{$\max_y p_{\bs{\theta}}(y|\bs{x})$} & Wide-ResNet             & \bf{.93} & \bf{.77} & .85 & .62 \\
                                      & Class-Conditional Glow  & .64 & .61 & .65 & .54 \\
                                      & IGEBM                   & .43 & .69 & .54 & .69 \\
                                      & JEM                     & .89 & .75 & \bf{.87} & \bf{.79} \\
                                      & GMMC (Gen)             & .84 & .72 & .81 & .31 \\
\midrule
\multirow{5}{*}{$\| \frac{\partial \log p_{\bs{\theta}}(\bs{x})}{\partial \bs{x}}\|$} & Unconditional Glow     & \bf{.95} & .27 & .46 & .29 \\
                            & Class-Conditional Glow & .47 & .01 & .52 & .59 \\
                            & IGEBM                  & .84 & .65 & .55 & .66 \\
                            & JEM                    & .83 & .78 & .82 & .79 \\
                            & GMMC (Gen)             & .88 & \bf{.79} & \bf{.85} & \bf{.87}\\
\bottomrule
\end{tabular}
\end{threeparttable}
\end{center}

\end{table*}

\begin{table*}[ht!]
  \centering
  \caption{Histograms of $\log_{\bs{\theta}}p(\bs{x})$ for OOD detection. Green corresponds to in-distribution dataset, while red corresponds to OOD dataset.}
  \begin{tabular}{ | c | m{2.9cm} | m{2.9cm} | m{2.9cm} | }
    \hline
    JEM
    &
    \begin{minipage}{.24\textwidth}
      \includegraphics[width=\linewidth, height=30mm]{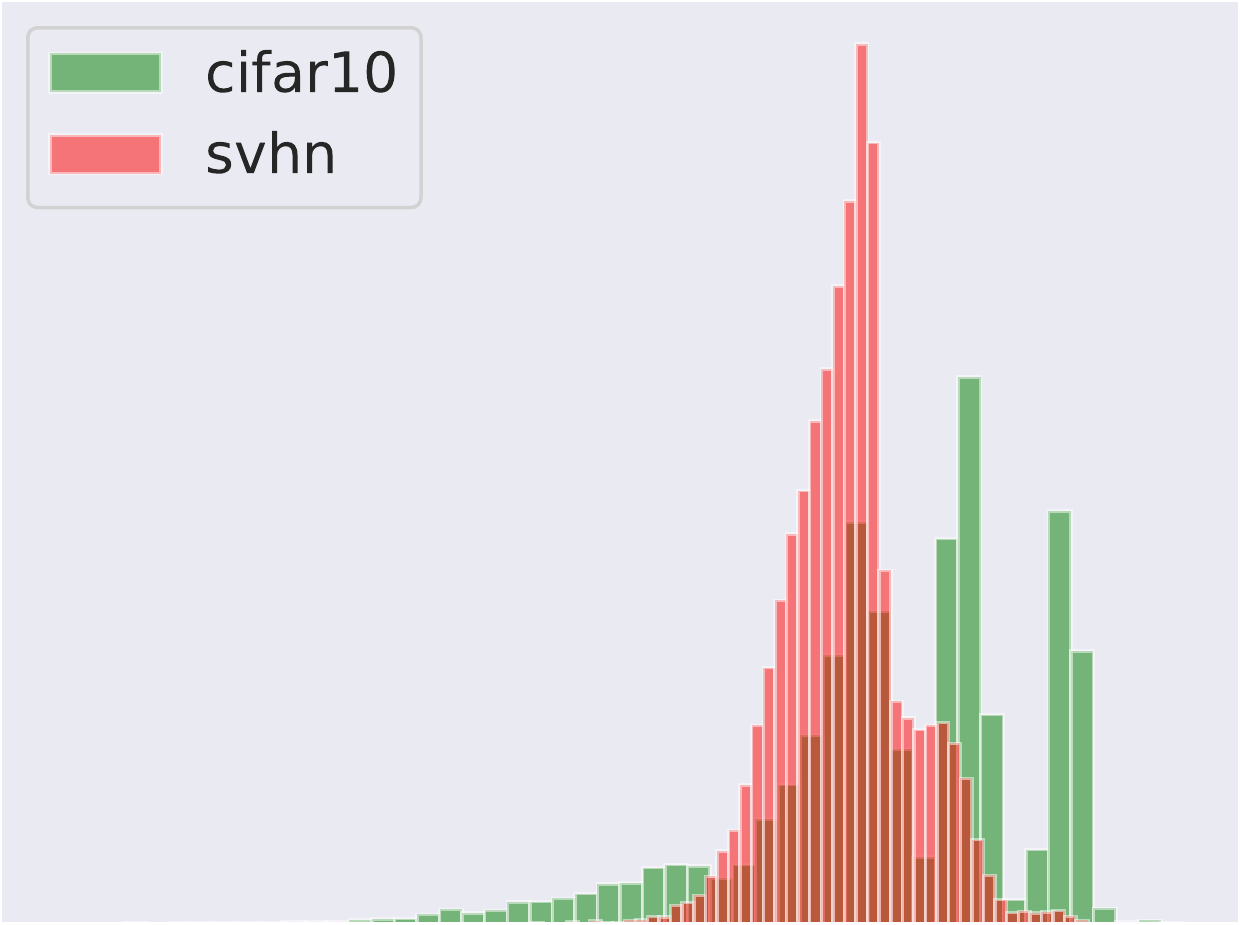}
    \end{minipage}
    &
    \begin{minipage}{.24\textwidth}
      \includegraphics[width=\linewidth, height=30mm]{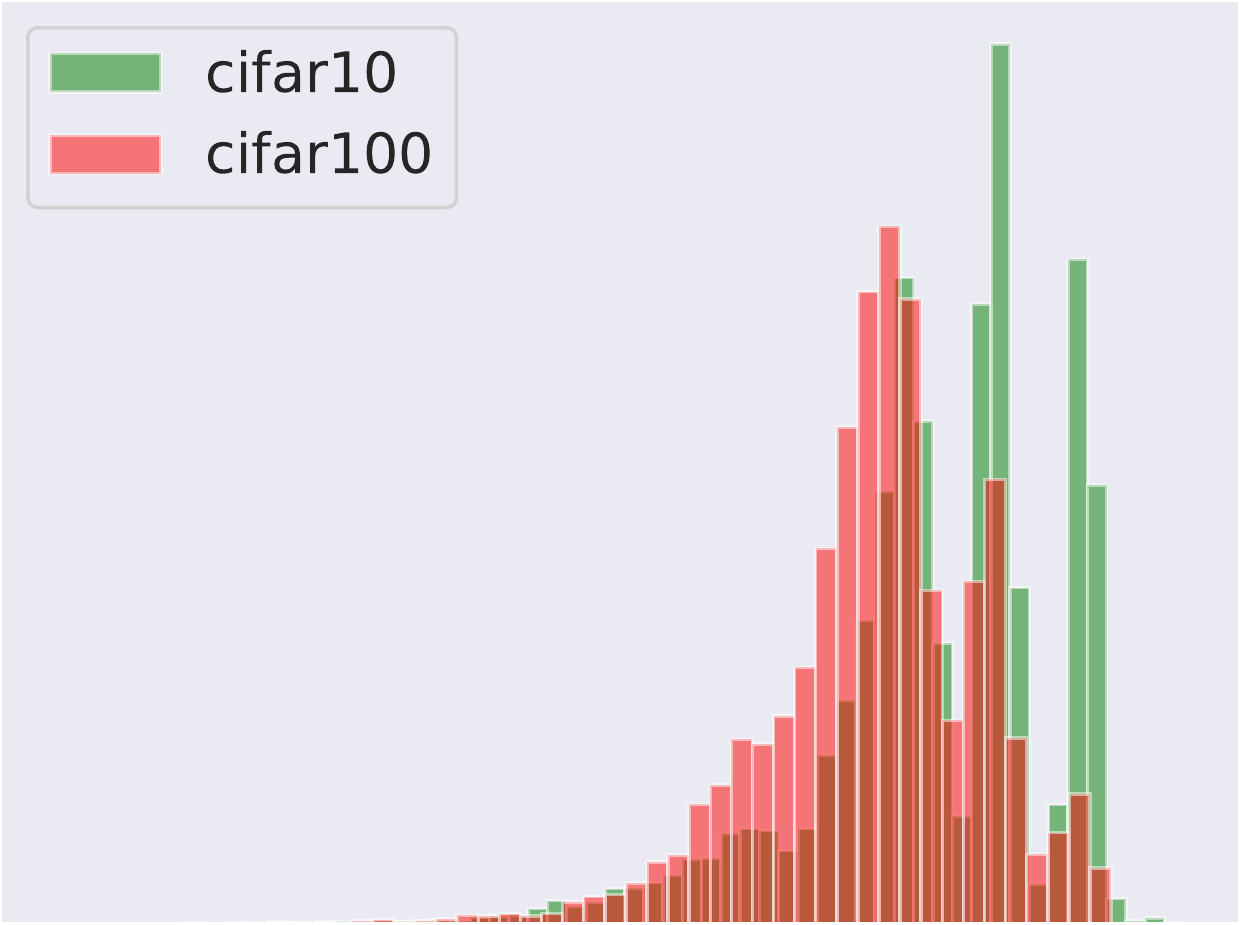}
    \end{minipage}
    &
    \begin{minipage}{.24\textwidth}
      \includegraphics[width=\linewidth, height=30mm]{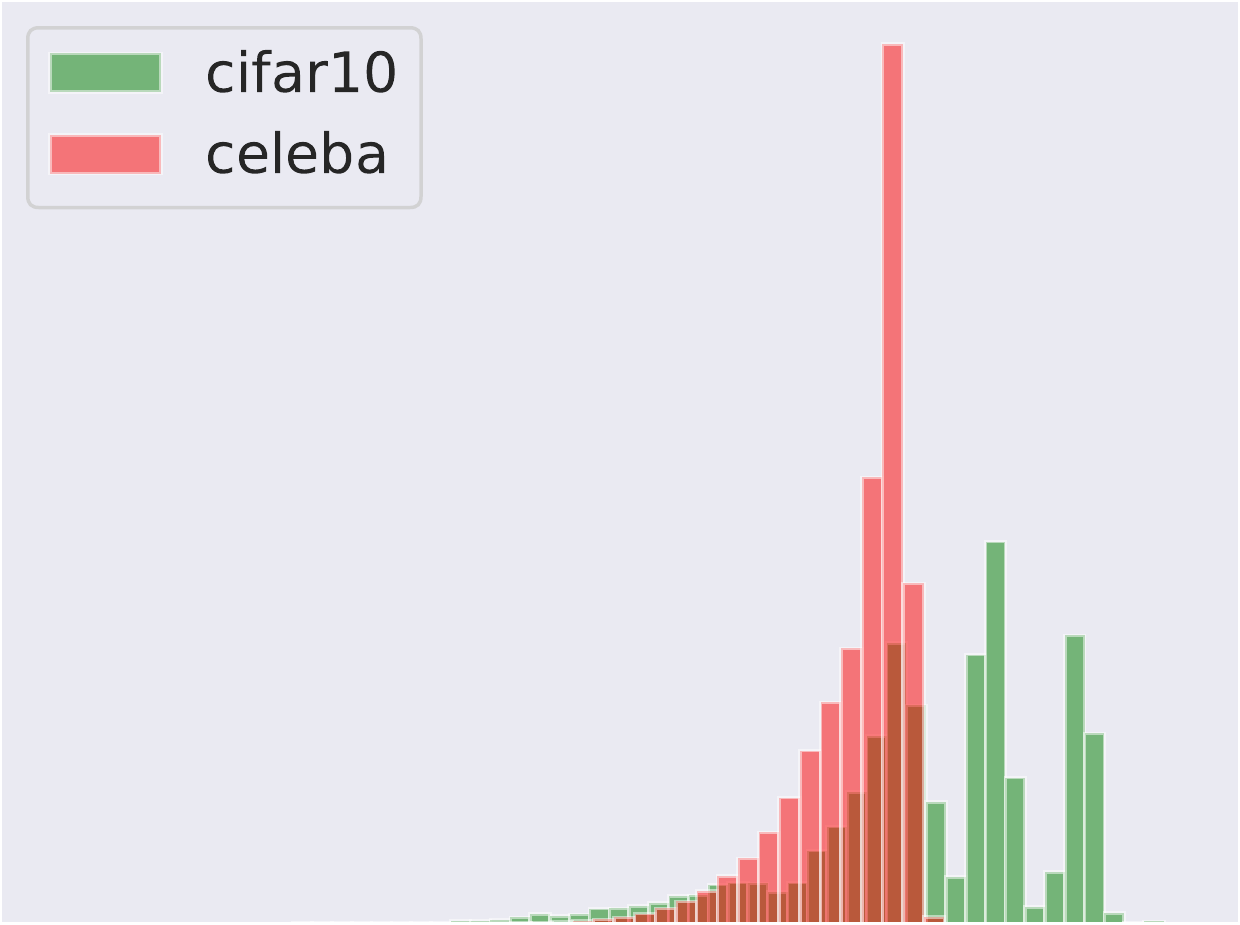}
    \end{minipage}
      \\ \hline
    GMMC (Gen)
    &
    \begin{minipage}{.24\textwidth}
      \includegraphics[width=\linewidth, height=30mm]{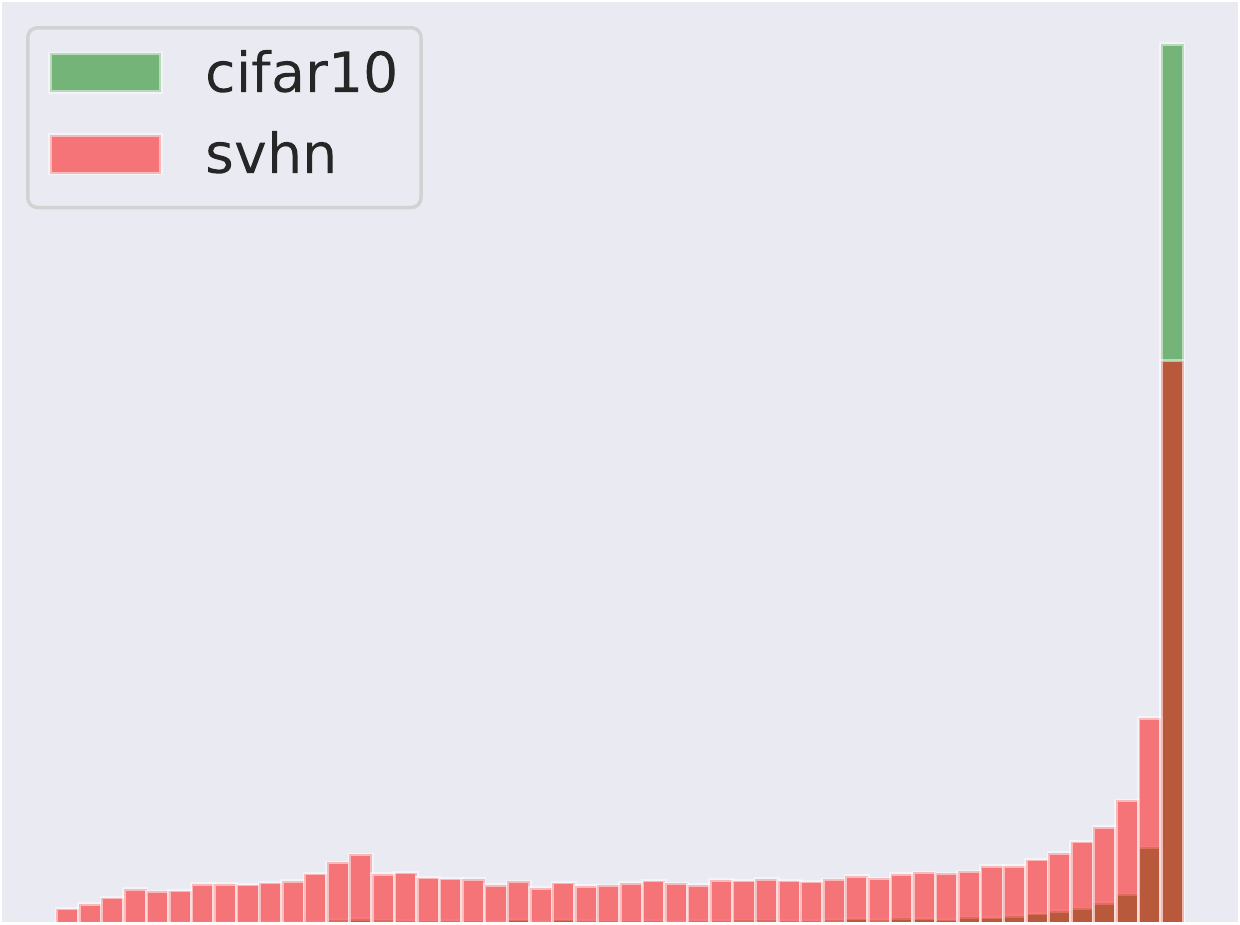}
    \end{minipage}
    &
    \begin{minipage}{.24\textwidth}
      \includegraphics[width=\linewidth, height=30mm]{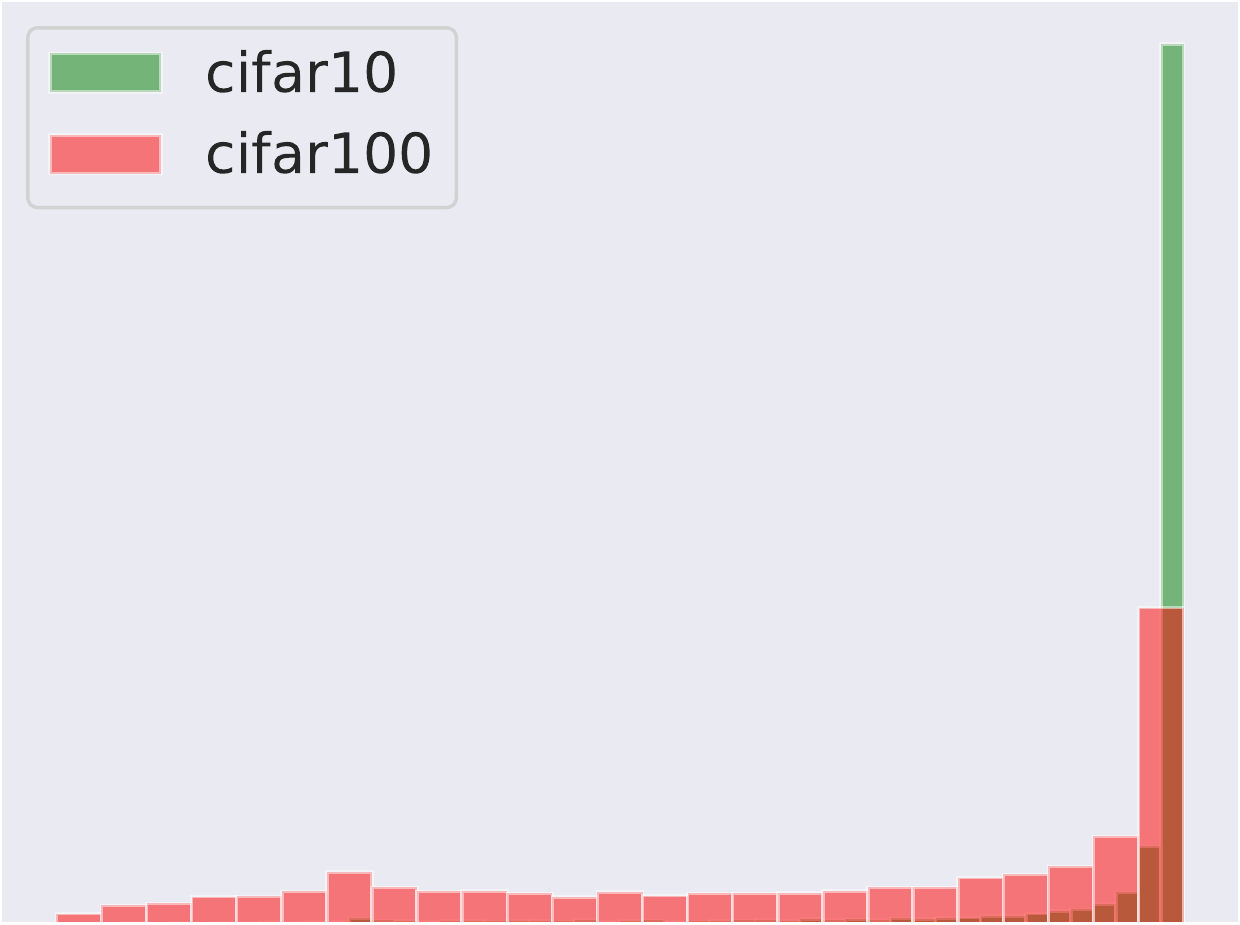}
    \end{minipage}
    &
    \begin{minipage}{.24\textwidth}
      \includegraphics[width=\linewidth, height=30mm]{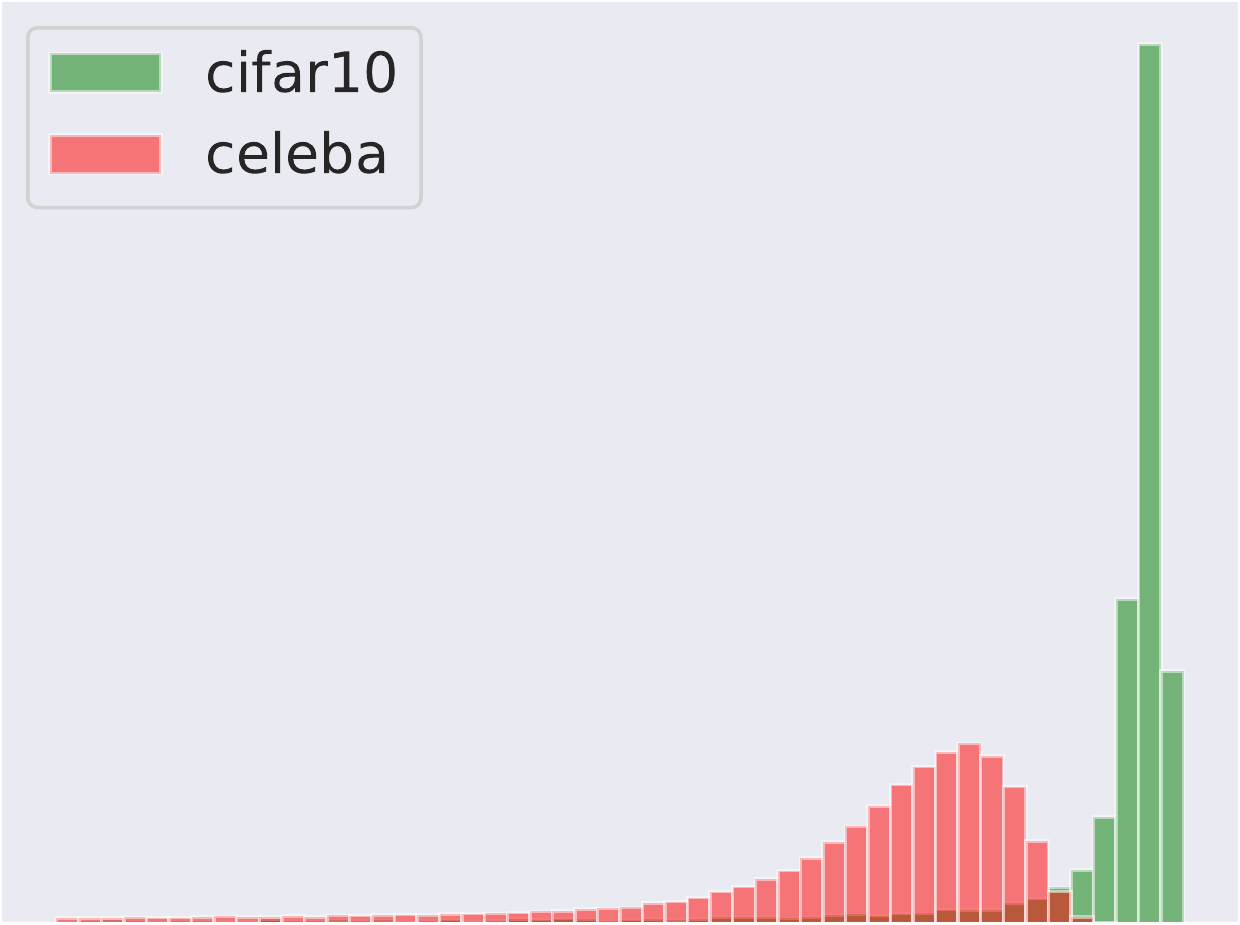}
    \end{minipage}
    \\ \hline
  \end{tabular}
  \label{table:logpx_hist}

\end{table*}

In summary, among all three different OOD score functions, GMMC outperforms the competing methods by a notable margin with two of them, while being largely on par with the rest one. The improved performance of GMMC on OOD detection is likely due to its \emph{explicit} generative modeling of $p_{\bs{\theta}}(\bs{x},y)$, which improves the evaluation of $p_{\theta}(\bs{x})$ over other methods.

\subsection{Robustness}

DNNs have demonstrated remarkable success in solving complex prediction tasks. However, recent works~\cite{Szegedy2013,advexample15,kurakin2016adversarial} have shown that they are particularly vulnerable to adversarial examples, which are in the form of small perturbations to inputs but can lead DNNs to predict incorrect outputs. Adversarial examples are commonly generated through an iterative optimization procedure, which resembles the iterative sampling procedure of SGLD in Eq.~\ref{eq:sgld}. GMMC (and JEM) further utilizes sampled data along with real data for model training (see Eq.~\ref{eq:ebm_grad}). This again  shares some similarity with adversarial training~\cite{advexample15}, which has been proved to be the most effective method for adversarial defense. In this section, we show that GMMC achieves considerable robustness compared to other methods thanks to the MMC modeling~\cite{Pang2020Rethinking} and its generative training.

\begin{table*}[ht!]
\caption{Classification accuracies when models are under $L_\infty$ PGD attack with different $\epsilon$'s. All models are trained on CIFAR10.}\label{table:cifar10_pgd_inf}\vspace{-5pt}
\begin{center}
\begin{threeparttable}
\begin{tabular}{c|c|cccc}
\toprule
\multirow{2}{*}{Model}     & \multirow{2}{*}{Clean (\%)}  & PGD-40 & PGD-40 & PGD-40 & PGD-40 \\
& & $\epsilon=4/255$ & $\epsilon=8/255 $ & $\epsilon=16/255$ & $\epsilon=32/255 $\\
\midrule
Softmax                 & 93.56  & 19.05 &  4.95 &  0.57 &  0.06 \\
JEM                     & 92.83  & 34.39 &  21.23 &  5.82 & 0.50 \\
GMMC (Dis)              & \bf{94.34} & 44.83 & 42.22 & 41.76 & 38.69\\
GMMC (Gen)              & 94.13  & \bf{56.29} & \bf{56.27} & \bf{56.13} & \bf{55.02} \\
\bottomrule
\end{tabular}
\end{threeparttable}
\end{center}

\end{table*}

\paragraph{\textbf{(1) PGD Attack}}
We run the white-box PGD attack~\cite{madry2018towards} on the models trained by standard softmax, JEM and GMMC. We use the same number of steps (40) with different $\epsilon$'s as the settings of JEM for PGD. Table~\ref{table:cifar10_pgd_inf} reports the test accuracies of different methods. It can be observed that GMMC achieves much higher accuracies than standard softmax and JEM under all different attack strengths. The superior defense performance of GMMC (Dis) over JEM mainly attributes to the MMC modeling~\cite{Pang2020Rethinking}, while GMMC (Gen) further improves its robustness over GMMC (Dis) due to the generative training.

\paragraph{\textbf{(2) C\&W Attack}}
Pang et al.~\cite{pang2018max} reveal that when applying the C\&W attack~\cite{CW_attack} on their trained networks some adversarial noises have clearly interpretable semantic meanings to the original images. Tsipras et al.~\cite{tsipras2018robustness} also discover that the loss gradient of adversarially trained robust model aligns well with human perception. Interestingly, we observe similar phenomenon from our GMMC model. Fig.~\ref{figure:adv_noise} shows some examples of adversarial noises generated from the GMMC (Gen) model under the C\&W attack, where the noises are calculated as $(\bs{x}_{adv} - \bs{x}) / 2$ to keep the pixel values in $[-0.5, 0.5]$. We observe around 5\% of adversarial noises have clearly interpretable semantic meanings to their original images. These interpretable adversarial noises indicate that GMMC (Gen) can learn robust features such that the adversarial examples found by the C\&W attack have to weaken the features of the original images as a whole, rather than generating salt-and-pepper like perturbations as for models of lower robustness.

To have a quantitative measure of model robustness under C\&W attack, we apply the C\&W attack to the models trained by the standard softmax, JEM and GMMC. In terms of classification accuracy, all of them achieve almost 100\% error rate under the C\&W attack. However, as shown in Table~\ref{table:cw_attack}, the adversarial noises to attack the GMMC (Gen) model have a much larger $L_2$ norm than that of adversarial noises for other models. This indicates that to attack GMMC (Gen), the C\&W attack has to add much stronger noises in order to successfully evade the network. In addition, GMMC (Dis) achieves a similar robustness as JEM under the C\&W attack, while GMMC (Gen) achieves an improved robustness over GMMC (Dis) likely due to the generative training of GMMC.

\begin{table}[H]
\begin{minipage}{0.48\linewidth}

    \centering
    \includegraphics[width=0.85\columnwidth]{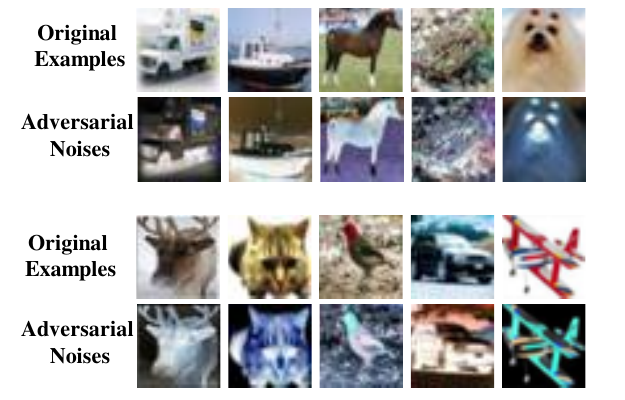}
    \caption{Example adversarial noises generated from the GMMC (Gen) model under C\&W attack on CIFAR10.}
    \label{figure:adv_noise}
\end{minipage}
\begin{minipage}{0.48\linewidth}

\caption{$L_2$ norms of adversarial perturbations under C\&W attack on CIFAR10.}\label{table:cw_attack}\vspace{-5pt}
\begin{center}
\begin{threeparttable}
\begin{tabular}{c|cc}
\toprule
\multirow{2}{*}{Model}  & Untarget    & Target   \\
                        &  iter=100   &  iter=1000 \\
\midrule
Softmax        & 0.205  & 0.331 \\
JEM            & 0.514  & 0.905 \\
GMMC (Dis)  & 0.564  & 0.784 \\
GMMC (Gen)  & \bf{1.686}  & \bf{1.546} \\
\bottomrule
\end{tabular}
\end{threeparttable}
\end{center}

\end{minipage}

\end{table}

\subsection{Training Stability}

Compared to JEM, GMMC has a well-formulated latent feature distribution, which fits well with the generative process of image synthesis. One advantage we observed from our experiments is that GMMC alleviates most of the instability issues of JEM. Empirically, we find that JEM can train less than 60 epochs before running into numerical issues, while GMMC can run 150 epochs smoothly without any numerical issues in most of our experiments.

\subsection{Joint Training}

Finally, we compare the joint training of JEM and GMMC on CIFAR10. The results show that joint training of GMMC is quite stable in most of our experiments, while JEM experiences substantial numerical instability issues. However, the quality of generated images from joint training of GMMC is not as good as generative training of GMMC from scratch. Due to page limit, details of the comparison are relegated to  Appendix~\ref{app:joint}.

\section{Conclusion and Future work}

In this paper, we propose GMMC by reinterpreting the max-Mahalanobis classifier~\cite{Pang2020Rethinking} as an EBM. Compared to the standard softmax classifier utilized in JEM, GMMC models the latent feature space explicitly as the max-Mahalanobis distribution, which aligns well with the generative process of image synthesis. We show that GMMC can be trained discriminative, generatively or jointly with reduced complexity and improved stability compared to JEM. Extensive experiments on the benchmark datasets demonstrate that GMMC can achieve state-of-the-art discriminative and generative performances, and improve calibration, out-of-distribution detection and adversarial robustness.

As for future work, we plan to investigate the GMMC models trained by different methods: discriminative vs. generative. We are interested in the differences between the features learned by different methods. We also plan to investigate the joint training of GMMC to improve the quality of generated images further because joint training speeds up the learning of GMMC significantly and can scale up GMMC to large-scale benchmarks, such as ImageNet.

\vspace{-10px}
\section{Acknowledgment}
\vspace{-5px}
We would like to thank the anonymous reviewers for their comments and suggestions, which helped improve the quality of this paper. We would also gratefully acknowledge the support of VMware Inc. for its university research fund to this research.
\vspace{-5px}

%
%
%
\bibliographystyle{splncs04}

\bibliography{ml}

\appendix

\section{Image Classification Benchmarks}\label{app:datasets}
The three image classification benchmarks used in our experiments are described below:
\begin{enumerate}
\item CIFAR10~\cite{Krizhevsky2012} contains 60,000 RGB images of size $32\times 32$ from 10 classes, in which 50,000 images are for training and 10,000 images are for test.
\item CIFAR100~\cite{Krizhevsky2012} also contains 60,000 RGB images of size $32\times 32$, except that it contains 100 classes with 500 training images and 100 test images per class.
\item SVHN~\cite{svhn11} is a street view house number dataset containing 73,257 training images and 26,032 test images classified into 10 classes representing digits. Each image may contain multiple real-world house number digits, and the task is to classify the center-most digit. 
\end{enumerate}

\section{Experimental Details}\label{app:exp}

Following JEM~\cite{jem}, all our experiments are based on the Wide-ResNet architecture~\cite{wideresnet16}. To ensure a fair comparison, we follow the same configurations of JEM by removing batch normalization and dropout from the network. We use ADAM optimizer~\cite{adam2014} with the initial learning rate of 0.0001 and the decay rate of 0.3, and train all our models for 150 epochs. For CIFAR10 and SVHN, we reduce the learning rate at epoch [30, 50, 80], while for CIFAR100 we reduce the learning rate much later at epoch [120, 140]. The hyperparameters of our generative training of GMMC are listed in Table~\ref{table:hyperparameters}. Compared to the configurations of IGEBM~\cite{du2019implicit} and JEM, we use a 10x larger buffer size and set the reinitialization frequency to 2.5\%. We note that with these settings GMMC generates images of higher quality. Comparing the two sampling approaches: Staged Sampling and Noise Injected Sampling, we find that their performances are very similar. Therefore, only the performances of Staged Sampling are reported.

\begin{table}[ht!]
\caption{Hyperparameters for generative training of GMMC}
\label{table:hyperparameters}\vspace{-10pt}
\begin{center}
\begin{threeparttable}
\begin{tabular}{l|cc}
\toprule
Variable      & Value \\
\midrule
Balance factor $\beta$   & 0.5        \\
Number of steps $\tau$       & 20         \\
Buffer size $|B|$         & 100,000      \\
Reinitialization freq. $\rho$  & 2.5\% \\
Step-size $\alpha$  & 1.0          \\
\bottomrule
\end{tabular}
\end{threeparttable}
\end{center}
\end{table}

\section{Pre-designed $\mu$ of MMC}
\label{app:mu}
Algorithm~\ref{algo:mmc_init} describes the construction of a set of means $\bs{\mu}=\{\bs{\mu}_y, y=1,2,\cdots,C\}$ that satisfy the condition of Max-Mahalanobis Distribution (MMD)~\cite{pang2018max}. $C$ is the number of classes, $d$ is the dimension of feature vector $\bs{\phi}(x)\in\mathbb{R}^d$ extracted by CNN, and $S$ is a hyperparameter, which we set to 10 in all our experiments. For more details of MMD, please refer to~\cite{pang2018max}.

\begin{algorithm}[h]
\caption{GenerateOptMeans($C,d,S$) for MMC~\cite{pang2018max}}
\label{algo:mmc_init}
\begin{algorithmic}
\STATE {\bfseries Input:} Number of classes $C$, dimension of feature vector $d$, and hyperparameter $S$. ($C\leq d+1$)\\
\STATE {\bfseries Initialization:} Let the $C$ mean vectors be $\bs{\mu}_1^*=e_1$ and $\bs{\mu}_i^*=0_{d},i\neq 1$. Here $e_1$ and $0_{d}$ denote the first unit basis vector and the zero vector in $\R^{d}$, respectively.
\FOR{$i=2$ {\bfseries to} $C$}
\FOR{$j=1$ {\bfseries to} $i-1$}
\STATE $\bs{\mu}^*_i(j)=-[1+\langle \bs{\mu}^*_i, \bs{\mu}^*_j \rangle \cdot (C-1)]/[ \bs{\mu}^*_j(j) \cdot(C-1)]$
\ENDFOR
\STATE $\bs{\mu}^*_i(i)=\sqrt{1-\lVert\bs{\mu}^*_i \rVert^2_2}$
\ENDFOR
 \FOR{$k=1$ {\bfseries to} $C$}
\STATE $\bs{\mu}^*_k=S\cdot \bs{\mu}^*_k$
\ENDFOR
\STATE {\bfseries Return:} The optimal mean vectors $\bs{\mu}^*_i,i\in[C]$.
\end{algorithmic}
\end{algorithm}



\section{Calibration}
\paragraph{Expected Calibration Error} (ECE) is a commonly adopted metric to measure the calibration of a model. First, it computes the confidence of the model, $\max_y p(y|\bs{x}_i)$, for each $\bs{x}_i$ in the dataset. Then it groups the predictions into equally spaced buckets $\{B_1,B_2,\cdots, B_M\}$ based on their confidence scores. For example, if $M$ = 20, then $B_1$ would represent all examples for which the model's confidence scores were between 0 and 0.05. Then ECE is calculated as
\begin{equation}
    \mathrm{ECE}=\sum_{m=1}^{M} \frac{\left|B_{m}\right|}{n}\left|\operatorname{acc}\left(B_{m}\right)-\operatorname{conf}\left(B_{m}\right)\right|,
\end{equation}
where $n$ is the number of examples in the dataset, acc($B_m$) is the average accuracy of the model on all the examples in $B_m$ and conf($B_m$) is the average confidence on all the examples in $B_m$. In our experiments, we set $M$ = 20. For a perfectly calibrated model, the ECE will be 0 for any $M$.

\subsection{Calibration Results on CIFAR100 and SVHN}\label{app:calibration}

\begin{figure}[ht!]
    \centering
    \subfigure[SVHN]{
        \includegraphics[width=0.41\columnwidth]{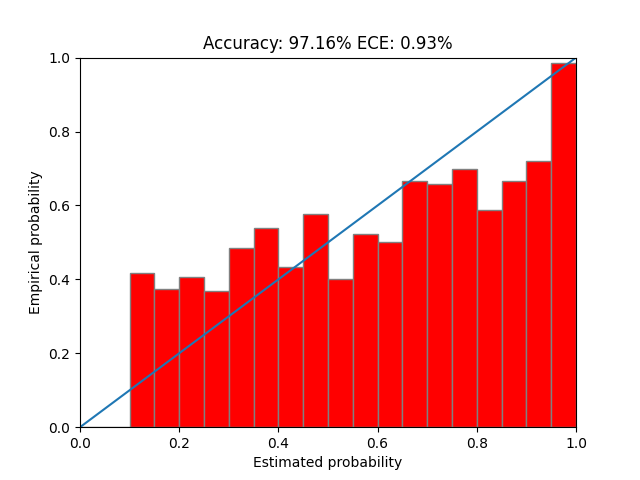}
        \label{figure:svhn_cali}
    }
    \subfigure[CIFAR100]{
        \includegraphics[width=0.41\columnwidth]{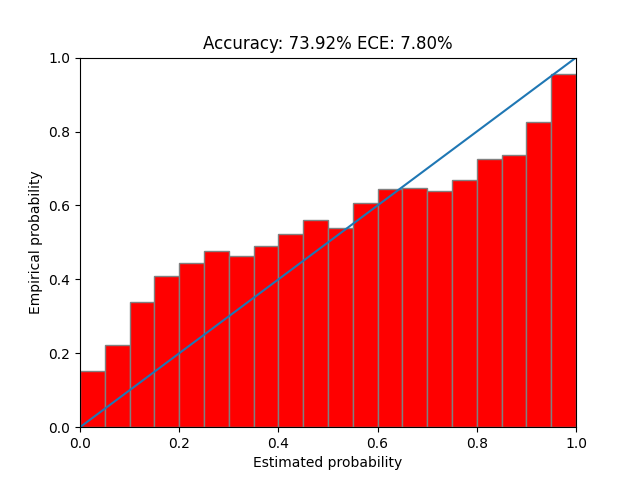}
        \label{figure:cifar100_cali}
    }\vspace{-10pt}
    \caption{GMMC calibration results on SVHN and CIFAR100.}
    \label{figure:svhn_cifar100_cali}
\end{figure}

Fig.~\ref{figure:svhn_cifar100_cali} and Table~\ref{table:cali_svhn_cifar100} show the calibration results of GMMC on SVHN and CIFAR100. As we can see, GMMC outperforms the softmax baseline and JEM on SVHN significantly, while JEM works better on CIFAR100.

\vspace{-20pt}
\begin{table}[ht!]
\caption{Calibration results on SVHN and CIFAR100.}
\label{table:cali_svhn_cifar100}\vspace{-10pt}
\begin{center}
\begin{threeparttable}
\begin{tabular}{c|cc}
\toprule
Model          &   SVHN  &  CIFAR100 \\
\midrule
Softmax        &  1.91  & 22.32 \\
JEM            &  2.08  &  \bf{4.87} \\
GMMC           &  \bf{0.93}  &  7.80 \\
\bottomrule
\end{tabular}
\end{threeparttable}
\end{center}
\end{table}



\vspace{-30pt}
\section{Joint Training}\label{app:joint}

Comparing Eq.~5 and Eq.~7 in the main text, we note that the gradient of Eq.~5 is just the second term of Eq.~7. Hence, we can use (Approach 1) discriminative training to pretrain $\bs{\theta}$, and then finetune $\bs{\theta}$ by (Approach 2) generative training. The transition between the two can be achieved by scaling up $\beta$ from 0 to a predefined value (e.g., 0.5). Similar joint training strategy could be applied to train JEM. However, from our experiments we note that this joint training of JEM is extremely unstable.

\begin{figure}[ht!]
    \centering
    \includegraphics[width=0.8\columnwidth]{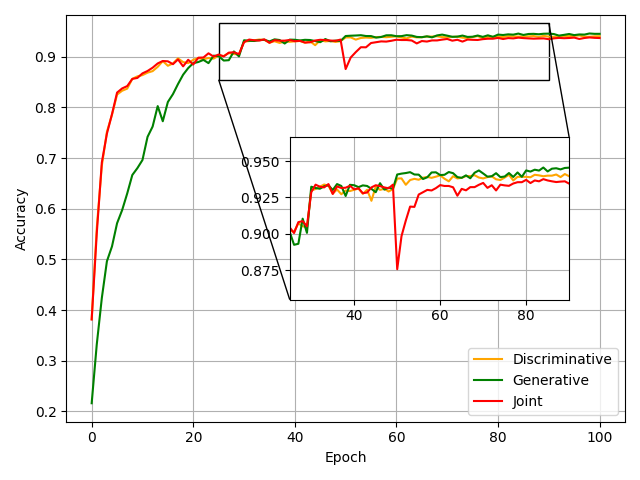}\vspace{-15pt}
    \caption{Validation accuracy curves of GMMC trained with discriminative, generative or joint training.}
    \label{figure:cifar10_transfer_curves}
\end{figure}

\begin{figure}[ht!]
    \centering
    \includegraphics[width=0.5\columnwidth]{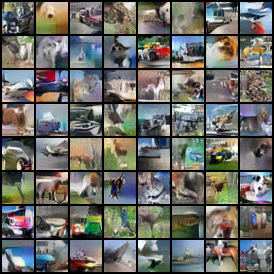}\vspace{-5pt}
    \caption{Generated CIFAR10 samples (unconditional) by GMMC with joint training. Compared to the samples from GMMC with generative training, these samples are more blurry.}
    \label{figure:cifar10_transfer}
\end{figure}

Fig.~\ref{figure:cifar10_transfer_curves} shows the validation accuracy curves of GMMC trained with discriminative, generative and joint training on CIFAR10. We train each model with 100 epochs, while for the joint training the first 50 epochs are with discriminative training and the rest of 50 epochs are switched to generative training. As we can see, all three models achieve a very similar accuracy of 93.5\% in the end. For the joint training, its accuracy drops a little bit around the switching epoch (51th), but finally can catch up to achieve a similar accuracy of 93.5\%. Interestingly, we note that the quality of sampled images from joint training, as shown in Fig.~\ref{figure:cifar10_transfer}, is not as good as that of generative training from scratch. Nevertheless, this results demonstrate that we can potentially use the joint training to speed up the training of GMMC, while generating competitive samples. We leave the improvement of image quality of the joint training of GMMC to future work. 

\vspace{-5pt}

\vspace{-25pt}
\vspace{-10pt}


Similar to the joint training of GMMC, we explore the joint training of JEM, i.e., pretrain the softmax classifier with standard distriminative training and then finetune the softmax classifier with JEM. Table~\ref{table:transfer_jem} shows the evolution of energy scores of real data and sampled data before and after the switching epoch. Here the energy function is defined as $E_{\bs{\theta}}(\bs{x}) = -\log \sum_y \exp(f_{\bs{\theta}}(\bs{x})[y])$ according to JEM. As can be seen, the energy scores on real data and sampled data are quickly exploded within a few iterations, demonstrating the instability of joint training of JEM. We also use different $\tau$s, the number of SGLD steps, to stabilize JEM but with no success.

\begin{table*}[ht!]
\caption{The energy scores of real data and sampled data after switching from discriminative training to JEM on CIFAR10.}\vspace{-5pt}

\label{table:transfer_jem}
\begin{center}
\begin{threeparttable}
\begin{tabular}{c|ccc|ccc}
\toprule
\multirow{2}{*}{Iteration}  & \multicolumn{3}{c}{Real Data}  & \multicolumn{3}{c}{Sampled Data}  \\
 & $\tau=20$ & $\tau=50$ & $\tau=100$ & $\tau=20$ & $\tau=50$ & $\tau=100$ \\
\midrule
Before transfer   & -22.4 & -21.2 & -21.6 & \multicolumn{3}{c}{N/A}  \\
1      &  -21.5 & -19.3 & -20.7   &  -2752 &  -7369 & -13412 \\
2      &  -9.7 & -7.6 & -8.5      &  -67  & -243 & -556 \\
3      &  -5.4 & -3.9 & -3.7      &  -4 & -86 & -449 \\
4      &  -3.2 & -2.2 & -0.8      &  294 & 695 & 1711\\
5      &  -15.4 & 10.5 & 13.9 &  4309 & 3182 & 6364\\
6      &  524 & 693 & 810   &  1.04e+7 & 6.34e+6 & 1.01e+7\\
7      &  18676 & 18250 & 23021 & 3.11e+9 & 3.01e+9 & 5.15e+9 \\
\bottomrule
\end{tabular}
\end{threeparttable}
\end{center}
\end{table*}

\vspace{-20pt}
\section{Additional GMMC Generated Samples}\label{app:samples}

\begin{figure}[ht!]
    \centering
    \subfigure[SVHN (Conditional)]{
        \includegraphics[width=0.41\columnwidth]{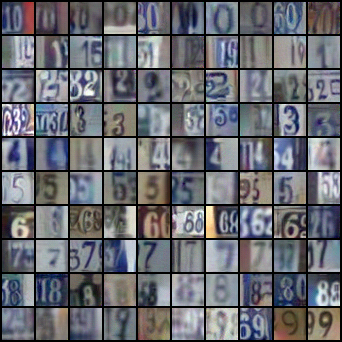}
        \label{figure:svhn_cond}
    }
    \subfigure[CIFAR100 (Conditional)]{
        \includegraphics[width=0.41\columnwidth]{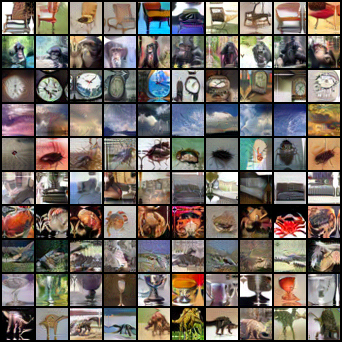}
        \label{figure:cifar100_cond}
    }\vspace{-5pt}
    \caption{GMMC generated class-conditional samples of SVHN and CIFAR100. Each row corresponds to one class.}
    \label{figure:svhn_cifar100_cond}
\end{figure}

\begin{figure*}[ht!]
    \centering
    \subfigure[Closest to the class center]{
        \includegraphics[width=0.41\columnwidth]{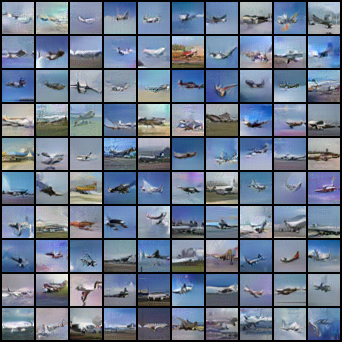}
    }
    \subfigure[Farthest to the class center]{
        \includegraphics[width=0.41\columnwidth]{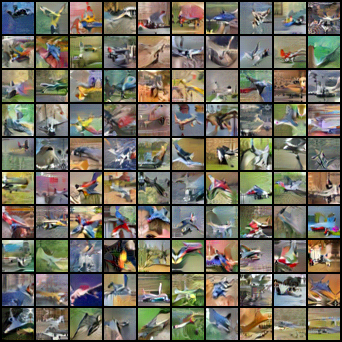}
    }\vspace{-5pt}
    \caption{GMMC generated class-conditional samples of \textbf{Plane}}
    \label{figure:class_0}
\end{figure*}

\begin{figure*}[ht!]
    \centering
    \subfigure[Closest to the class center]{
        \includegraphics[width=0.41\columnwidth]{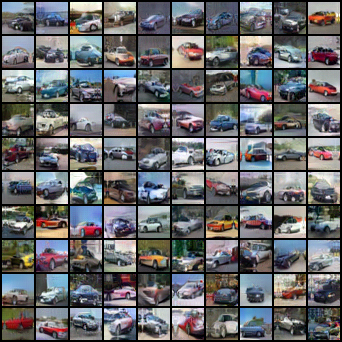}
    }
    \subfigure[Farthest to the class center]{
        \includegraphics[width=0.41\columnwidth]{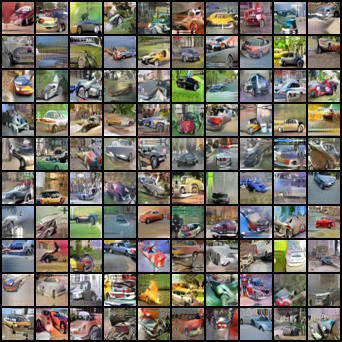}
    }\vspace{-5pt}
    \caption{GMMC generated class-conditional samples of \textbf{Car}}
    \label{figure:class_1}
\end{figure*}

\begin{figure*}[ht!]
    \centering
    \subfigure[Closest to the class center]{
        \includegraphics[width=0.41\columnwidth]{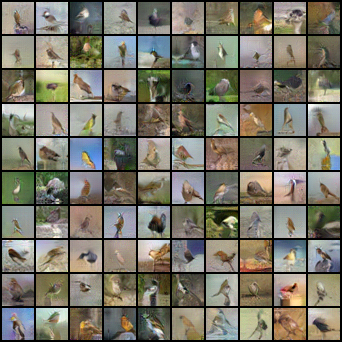}
    }
    \subfigure[Farthest to the class center]{
        \includegraphics[width=0.41\columnwidth]{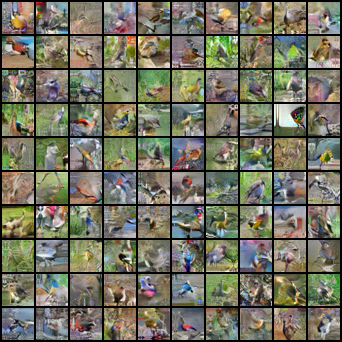}
    }\vspace{-5pt}
    \caption{GMMC generated class-conditional samples of \textbf{Bird}}
    \label{figure:class_2}
\end{figure*}

\begin{figure*}[ht!]
    \centering
    \subfigure[Closest to the class center]{
        \includegraphics[width=0.41\columnwidth]{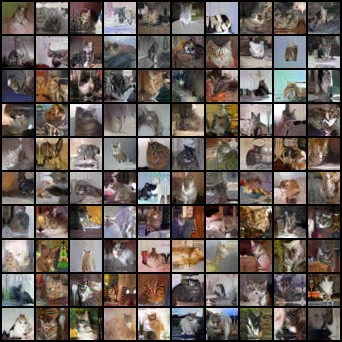}
    }
    \subfigure[Farthest to the class center]{
        \includegraphics[width=0.41\columnwidth]{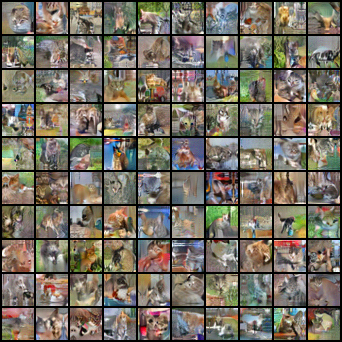}
    }\vspace{-5pt}
    \caption{GMMC generated class-conditional samples of \textbf{Cat}}
    \label{figure:class_3}
\end{figure*}

\begin{figure*}[ht!]
    \centering
    \subfigure[Closest to the class center]{
        \includegraphics[width=0.41\columnwidth]{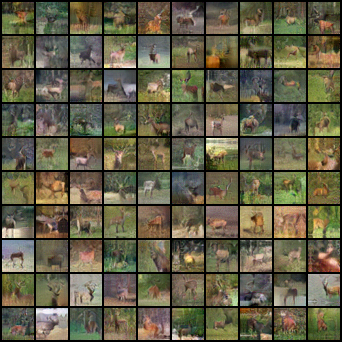}
    }
    \subfigure[Farthest to the class center]{
        \includegraphics[width=0.41\columnwidth]{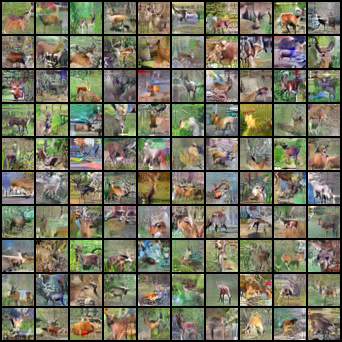}
    }\vspace{-5pt}
    \caption{GMMC generated class-conditional samples of \textbf{Deer}}
    \label{figure:class_4}
\end{figure*}

\begin{figure*}[ht!]
    \centering
    \subfigure[Closest to the class center]{
        \includegraphics[width=0.41\columnwidth]{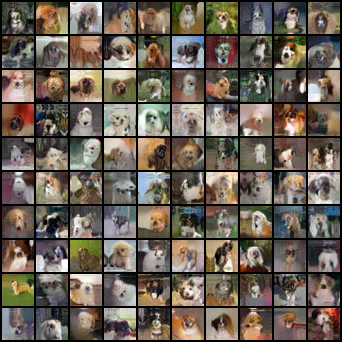}
    }
    \subfigure[Farthest to the class center]{
        \includegraphics[width=0.41\columnwidth]{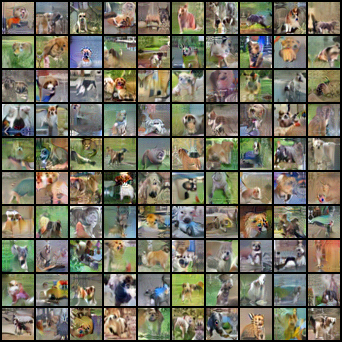}
    }\vspace{-5pt}
    \caption{GMMC generated class-conditional samples of \textbf{Dog}}
    \label{figure:class_5}
\end{figure*}

\begin{figure*}[ht!]
    \centering
    \subfigure[Closest to the class center]{
        \includegraphics[width=0.41\columnwidth]{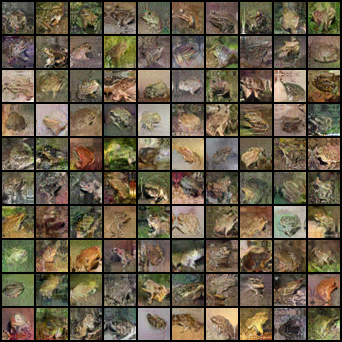}
    }
    \subfigure[Farthest to the class center]{
        \includegraphics[width=0.41\columnwidth]{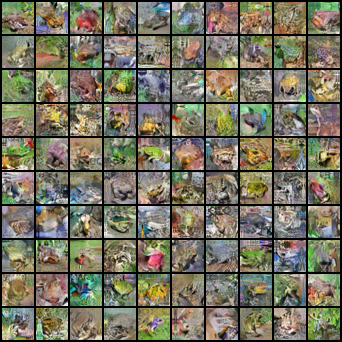}
    }\vspace{-5pt}
    \caption{GMMC generated class-conditional samples of \textbf{Frog}}
    \label{figure:class_6}
\end{figure*}

\begin{figure*}[ht!]
    \centering
    \subfigure[Closest to the class center]{
        \includegraphics[width=0.41\columnwidth]{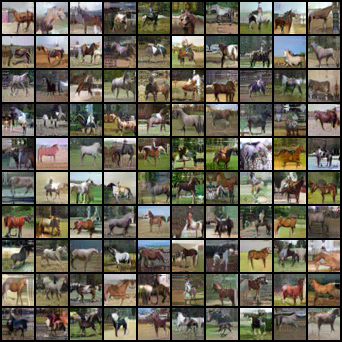}
    }
    \subfigure[Farthest to the class center]{
        \includegraphics[width=0.41\columnwidth]{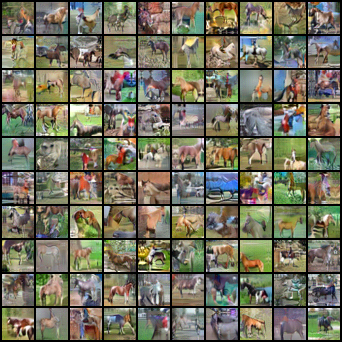}
    }\vspace{-5pt}
    \caption{GMMC generated class-conditional samples of \textbf{Horse}}
    \label{figure:class_7}
\end{figure*}

\begin{figure*}[ht!]
    \centering
    \subfigure[Closest to the class center]{
        \includegraphics[width=0.41\columnwidth]{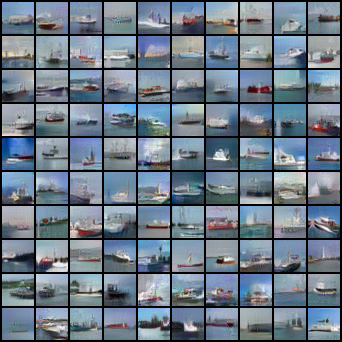}
    }
    \subfigure[Farthest to the class center]{
        \includegraphics[width=0.41\columnwidth]{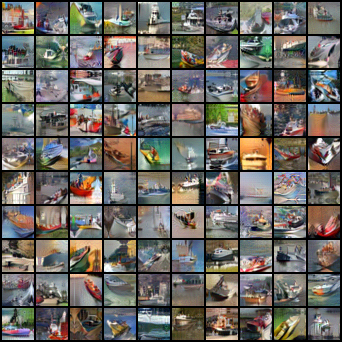}
    }\vspace{-5pt}
    \caption{GMMC generated class-conditional samples of \textbf{Ship}}
    \label{figure:class_8}
\end{figure*}

\begin{figure*}[ht!]
    \centering
    \subfigure[Closest to the class center]{
        \includegraphics[width=0.41\columnwidth]{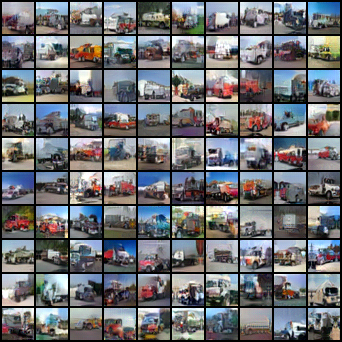}
    }
    \subfigure[Farthest to the class center]{
        \includegraphics[width=0.41\columnwidth]{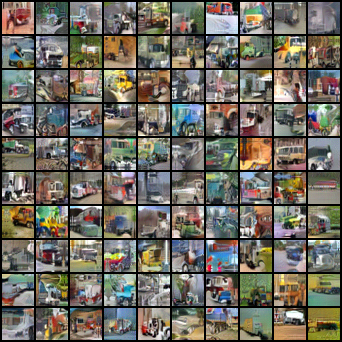}
    }\vspace{-5pt}
    \caption{GMMC generated class-conditional samples of \textbf{Truck}}
    \label{figure:class_9}
\end{figure*}

Additional GMMC generated samples of SVHN and CIFAR100 are provided in 
Fig.~\ref{figure:svhn_cifar100_cond}. GMMC generated class-conditional samples of CIFAR10 are provided in Figs~\ref{figure:class_0}-\ref{figure:class_9}. To evaluate the quality of generated images, we use the same IS and FID code as in IGEBM~\cite{du2019implicit} and JEM~\cite{jem} on the class-conditional samples.

\end{document}